\def\year{2021}\relax
\documentclass[letterpaper]{article} 
\usepackage{aaai21_modified}  
\usepackage{times}  
\usepackage{helvet} 
\usepackage{courier}  
\usepackage[hyphens]{url}  
\usepackage{graphicx} 
\usepackage{natbib}  
\usepackage{caption} 
\usepackage{times}

\usepackage{soul}
\usepackage{url}
\usepackage[hidelinks]{hyperref}
\usepackage[utf8]{inputenc}
\usepackage{placeins}
\usepackage{appendix}
\usepackage{amsmath}
\usepackage{mathtools}
\usepackage{booktabs}
\usepackage{amsthm}
\usepackage{subcaption}
\usepackage{multirow}
\usepackage{hyphenat}
\usepackage{amssymb}
\usepackage{tabularx}
\usepackage{xfrac}
\usepackage{todonotes}
\usepackage{enumitem}
\usepackage{adjustbox}
\usepackage{array}
\usepackage{tcolorbox}
\tcbuselibrary{breakable}

\newcolumntype{R}[2]{%
    >{\adjustbox{angle=#1,lap=\width-(#2)}\bgroup}%
    l%
    <{\egroup}%
}

\bgroup

\title{Online Action Recognition}

\author{ 
    Alejandro Su\'arez-Hern\'andez, \textsuperscript{\rm 1} 
    Javier Segovia-Aguas, \textsuperscript{\rm 1,2} 
    Carme Torras, \textsuperscript{\rm 1} 
    Guillem Aleny\`a\textsuperscript{\rm 1} \\
}

\affiliations{ 
\textsuperscript{\rm 1}  IRI - Institut de Robòtica i Informàtica Industrial, CSIC-UPC \\
\textsuperscript{\rm 2}  Universitat Pompeu Fabra \\
asuarez@iri.upc.edu, javier.segovia@upf.edu, torras@iri.upc.edu, galenya@iri.upc.edu
}

\graphicspath{{./figures/}}

\newtheorem{thm}{Theorem}

\newtheorem{example}{Example}
\newtheorem{definition}{Definition}

\newtheorem{condition}{Condition}

\newcommand{\strips}{\textsc{Strips}}

\newcommand{\thealgorithmacro}{OARU}
\newcommand{\thealgorithmfull}{Online Action Recognition through Unification}
\newcommand{\defeq}{\vcentcolon=}
\newcommand{\setstyle}[1]{\mathrm{#1}}
\newcommand{\typefn}[1]{\mathrm{type}(#1)}

\newcommand{\argfn}[2]{\mathrm{arg}_{#1}(#2)}

\usepackage{algorithm} 
\usepackage{algorithmic}

\begin{document}
\maketitle


\begin{abstract}
Recognition in planning seeks to find agent intentions, goals or activities given a set of observations and a knowledge library (e.g. goal states, plans or domain theories). 
In this work we introduce the problem of \textit{Online Action Recognition}. It consists in recognizing, in an {\em open world}, the planning action that \textit{best} explains a partially observable state transition from a knowledge library of first-order \strips{} actions, which is initially empty. We frame this as an optimization problem, and propose two algorithms to address it: \textit{Action Unification} (AU) and \textit{\thealgorithmfull} (\thealgorithmacro). The former builds on \textit{logic unification} and generalizes two input actions using \textit{weighted partial MaxSAT}. The latter looks for an action within the library that explains an observed transition. If there is such action, it generalizes it making use of AU, building in this way an AU hierarchy. Otherwise, \thealgorithmacro{} inserts a \textit{Trivial Grounded Action} (TGA) in the library that explains just that transition. We report results on benchmarks from the International Planning Competition and PDDLGym, where \thealgorithmacro{} recognizes actions accurately with respect to expert knowledge, and shows real-time performance.
\end{abstract}

\section{Introduction}
The prediction of the most likely actions, plans or goals of an agent, has been a topic of interest in the planning community since the work by \citeauthor{ramirez2009plan} (\citeyear{ramirez2009plan}), which posed the recognition task via planning given a domain theory and a set of observations. Other work adopted this convention of recognition as planning for related tasks, i.e. goal recognition and environment design \cite{keren2014goal}, production and recognition of context-free grammars \cite{segovia2017generating}, classification of planning instances in generalized plans \cite{segovia2017unsupervised},
counter-planning \cite{pozanco2018counterplanning}, model recognition \cite{aineto2019model} and recognition with noisy observations \cite{sohrabi2016plan,aineto2020observation}.

\begin{figure}[tb!]
    \centering
    \includegraphics[width=\columnwidth]{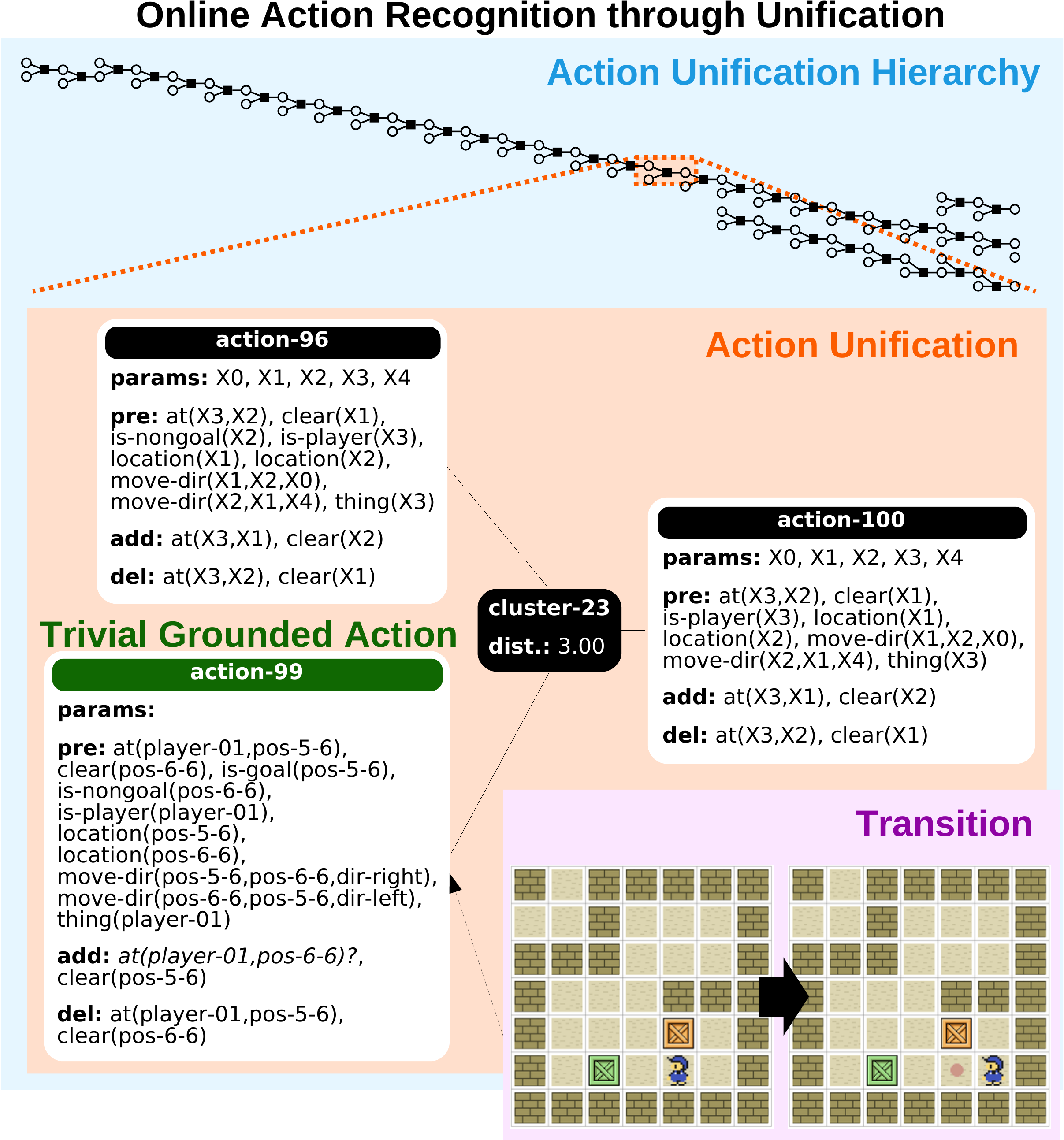}
    \caption{Illustration of AU and \thealgorithmacro{} in \textit{sokoban}.}
    \label{fig:teaser-figure}
\end{figure}


In this paper, we address the problem of {\em Online Action Recognition} (OAR), which consists in recognizing the actions performed by an agent in an {\em open-world}, given a {\em state transition} and a {\em knowledge library} of first-order \strips{} actions (also known as domain theory). We restrict OAR to symbolic inputs and deterministic action effects. However, we set an initially empty knowledge library that must be inductively filled from partially observable transitions. We describe an algorithm called {\em Action Unification} (AU), related to {\em syntactic logic unification}~\cite{snyder2001unification}. AU generalizes two actions using an encoding to {\em Weighted Partial MaxSAT} (WPMS). We also propose {\em \thealgorithmfull} (\thealgorithmacro{}), an algorithm that makes repeated use of AU to merge ad-hoc explanations of the transitions into its action library. In the following, we use \textit{action} as a generic category; \textit{schema} or \textit{model} for parameterized actions; \textit{grounded action} for parameter-less actions; and {\em Trivial Grounded Action} (TGA) for ad-hoc actions derived directly from observations.
\begin{example}
Figure~\ref{fig:teaser-figure} uses \textit{sokoban}~\cite{junghanns1997sokoban} to illustrate AU and \thealgorithmacro{}. The whole AU hierarchy, which summarizes the history of merged actions, is in the top. We zoom in one part, showing: (1) \texttt{action-96}, a schema already present in \thealgorithmacro{}'s library; (2) a TGA \texttt{action-99}, constructed from the transition depicted graphically at the bottom; and (3) a 5-ary schema \texttt{action-100} produced by AU, which generalizes the previous two actions. The transition corresponds to the character moving to the right. AU has determined that the {\em distance} between \texttt{action-96} and \texttt{action-99} is 3.
Later, we will see that this distance is calculated in terms of \textit{relaxed precondition predicates}.
Some predicates have a question mark indicating that they are uncertain due to the open world setting. We will also see how this uncertainty can be dispelled through AU.
Overall, the example shows how \thealgorithmacro{} surmised from the past that the player could not move from a goal cell, and corrects this assumption when shown otherwise.
\end{example}

The contributions of this work are: (1) a formal method for generalizing two planning actions with the AU algorithm, which deals with an \textit{NP-Hard} problem; (2) a scalable, accurate and suitable for real-time usage algorithm to recognize and acquire general schemata from partially observable state transitions, named \thealgorithmacro{}; and (3) an evaluation of acquired knowledge libraries with expert handcrafted benchmarks from the International Planning Competition (IPC)~\cite{muise2016planning.domains} and PDDLGym~\cite{Silver2020PDDLGym:Problems}.

\section{Related Work}
\label{sec:related-work}


The problem of OAR relates to plan, activity and intent recognition (PAIR) \cite{sukthankar2014plan} and learning action models \cite{arora2018review}. The recognition in PAIR is defined as a prediction task of the most plausible future, i.e. the most probable plan or goal an agent will pursue \cite{ramirez2009plan,ramirez2010probabilistic}, while our recognition task serves as an explanation of the past \cite{chakraborti2017plan,aineto2019model,aineto2020observation}. Also, model-based approaches for recognition assume a knowledge library with goals, plans or domain theories is known beforehand. We release the problem from this assumption, where the knowledge is acquired in an incremental online fashion.

We refer to learning the representation of the world dynamics as the problem of learning action models, which has been a topic of interest for long time \cite{gil1994learning,benson1995inductive,wang1995learning}. {\strips}-like actions have been learned with algorithms such as ARMS \cite{yang2007learning} with a weighted MaxSAT; SLAF \cite{amir2008learning} with an online SAT solver that computes the CNF formula compatible with partial observations; LOCM \cite{cresswell2009acquisition,cresswell2011generalised,gregory2015domain} computing finite state machines of object sorts; FAMA \cite{aineto2018learning,aineto2019learning} learning from minimal observability with an off-the-shelf classical planner; and Representation Discovery \cite{bonet2019learning} that learns with a SAT solver from plain graphs. In the latter, only action labels are known, while in the rest of approaches, the name and parameters of each declarative action are known, which strictly constraints the learning problem. In our case, no prior labels and parameters bound the problem, like in \citeauthor{suarez2020strips}~(\citeyear{suarez2020strips}), but inductively learned with every new partially observable state transition.

The work by \citeauthor{amado2018goal} (\citeyear{amado2018goal}) proposes the LatPlan algorithm for goal recognition. It first learns grounded action models in the latent space from images, such as AMA2 \cite{asai2018classical}. However, grounded actions are generated without soundness guarantees, may not generalize to other instances, and the learning method requires observing all possible non-symbolic transitions, while OARU inputs are symbolic, computes actions which do not require to observe all state transitions and generalize to other instances.

\section{Preliminaries}
Let us first introduce preliminary notation on first-order logic and unification, and weighted partial MaxSAT.


\subsection{First-order logic} 
The formal language to describe relations between constant symbols is known as {\em first-order logic} or {\em predicate logic}. {\em Propositional logic} is subsumed by {\em first-order logic} in that propositions are specific interpretations of relations over constant symbols. 
The {\bf syntax} of a first-order language consists of a (possibly) infinite set of {\em logical terms} (mathematical objects) and {\em well-formed formulae} (mathematical facts) denoted as $T$ and $\mathcal{F}$, respectively. 

\begin{definition}[First-Order Logical Term]
A {\em first-order logical term} $t \in T$ is recursively defined as $t = c | v | f_n( t_1, \ldots, t_n )$ where $t$ is either a constant symbol $c \in \mathcal{C}$, a variable symbol $v \in \mathcal{V}$, or a functional symbol $f_n$ with $n$ arguments s.t. $t_i \in T$ for all $1\leq i \leq n$. 
\end{definition}

\begin{definition}[Well-Formed Formula]
A {\em well-formed formula} (wff) $\varphi \in \mathcal{F}$ is inductively defined as a {\em predicate} (or atomic formula) $\varphi = p_n(t_1,\ldots,t_n)$; a {\em negation} $\neg \varphi$; a {\em logical connective}  conjunction $\varphi \wedge \psi$, disjunction $\varphi\vee\psi$, implication $\varphi \rightarrow \psi$, or biconditional $\varphi \leftrightarrow \psi$ s.t. $\psi$ is also a wff; or a {\em quantifier} $\forall_v \varphi$ or $\exists_v \varphi$, for universal and existential quantification of a wff $\varphi$ and variable $v\in{\cal V}$, respectively.
\end{definition}

The first-order {\bf semantics} is a structure which consists of an interpretation function $\mathcal{I}$ and a \textit{universe} $\mathcal{D}$ with the non-empty set of all objects. The set of objects is used to ascribe meaning to terms and formulae with an interpretation function but, while terms are interpreted into objects, predicates and other formulae have boolean interpretations.
The {\em interpretation over a logical term} $\mathcal{I}(t)$ is an assigned function to term $t$ which maps a tuple of already interpreted arguments of $n$ objects in $\mathcal{D}^n$ to a single object in $\mathcal{D}$. 
Hence, the interpretation of constant terms is $\mathcal{I}(c): \mathcal{D}^0 \rightarrow \mathcal{D}$, from variables is $\mathcal{I}(v): \mathcal{D}^1 \rightarrow \mathcal{D}$, and functional symbols is $\mathcal{I}(f): \mathcal{D}^n \rightarrow \mathcal{D}$, which maps $0$, $1$ and $n$ objects into one, respectively.
The interpretation of an $n$-ary predicate is $\mathcal{I}(p) : \mathcal{D}^n \rightarrow \{ \mathrm{true}, \mathrm{false} \}$.
The {\em interpretation over a well-formed formula} $\mathcal{I}(\varphi)$ consists of: (1) the interpretation of variables and functionals given a set of objects $\mathcal{D}$; and (2) the boolean evaluation of predicate interpretations, and logical connectives. 
In this paper we only focus on constant, variables and first-order predicate symbols such as follows:

\begin{example}
In Figure \ref{fig:teaser-figure}, there are 7 predicate symbols \{{\tt at}$_2$, {\tt clear}$_1$, {\tt is-player}$_1$, {\tt location}$_1$, {\tt move-dir}$_3$, {\tt thing}$_1$\}, 5 variables $\{X0,X1,X2,X3,X4\}$ and 5 constants \{{\tt player-01}, {\tt pos-5-6}, {\tt pos-6-6}, {\tt dir-left}, {\tt dir-right}\}. In the first state shown, the player is in position {\tt pos-5-6}, so $\mathcal{I}({\tt at}_2(\texttt{player-01},\texttt{pos-5-6})) = \mathrm{true} $. In the second one, it is no longer there, so the same interpretation is false. Predicates with variable arguments such as $ {\tt at}(X1,X0) $ require an interpretation of variables $ X1 $ and $ X0 $ before evaluation (known as grounding), e.g. $\{\mathcal{I}(X0)=\texttt{player-01}$, $\mathcal{I}(X1)=\texttt{pos-5-6}\}$.
\label{ex:first-order}
\end{example}

\subsection{First-Order Unification}

The problem of {first-order unification} ({\em unification} for short) is the algorithmic procedure to make both sides from a set of equations syntactically equal \cite{snyder2001unification}. Then, the unification problem is defined as a set of potential equations $\mathcal{U} = \{l_1 \doteq r_1, \ldots, l_n \doteq r_n\}$ where each left and right-hand sides are first-order logical terms $l_i,r_i \in T$. 

A {\em substitution} $\sigma : \mathcal{V} \rightarrow T$ is a function that maps variables into terms, and its notation is $\{v_1\rightarrow t_1, \ldots, v_n \rightarrow t_n\}$ where each $v_i \in \mathcal{V}$ and $t_i \in T$. An {\em instance} of a term $t$ is $t\sigma$ and expanded as $t\{v_1\rightarrow t_1, \ldots, v_n\rightarrow t_n\}$, where all variable substitutions are applied simultaneously. Two terms $l$ and $r$ of a potential equation $l\doteq r$ are syntactically equal if exists a substitution $\sigma$ such that $l\sigma \equiv r\sigma$, where $\equiv$ stands for syntactically equivalent.

\begin{example}[Substitution] Given a pair of constants $\mathit{player},\mathit{loc} \in \mathcal{C}$ and variables $v_1, v_2 \in \mathcal{V}$, the potential equation ${\tt at}(v_1,{\tt loc}) \doteq {\tt at}({\tt player},v_2)$ is syntactically equivalent when the substitution is $\sigma = \{ v_1 \rightarrow {\tt player}, v_2 \rightarrow {\tt loc} \}$.
\label{ex:substitution}
\end{example}


The substitution of terms is also named a {\em unifier}. A {\em solution} to $\mathcal{U}$ is a {\em unifier} $\sigma$ if $l_i\sigma \equiv r_i\sigma$ for all $1\leq i \leq n$. In the context of this paper, we are interested only in the unification of predicates without nested arguments (i.e. no functions).

\subsection{Weighted Partial MaxSAT (WPMS)} 
A {\em literal} $l$ is a boolean variable $l=x$ or its negation $l=\neg{x}$.
A {\em clause} $c$ is a disjunction of literals, and a {\em weighted clause} $(c,w)$ extends the clause $c$ with a natural number $w \in \mathbb{N}$ (or $\infty$) which represents the cost of not satisfying $(c,w)$. In WPMS, a weighted clause is either {\em hard} such that weights are infinite, forcing the solver to satisfy the clause, or {\em soft} where total accumulated weight of falsified soft clauses should be minimized (resp. maximize the ones that are satisfied).

In contrast to (partial Max) SAT where the input formula is described in conjunctive normal form (conjunctions of clauses), the input to a WPMS problem is a set of weighted clauses $\Phi = \{(c_1,w_1), \ldots, (c_n,w_n)\}$, which may be divided into $\Phi_s = \{(c_1,w_1),\ldots,(c_m,w_m)\}$ and $\Phi_h = \{(c_{m+1},\infty),\ldots,(c_n,\infty)\}$, for soft and hard weighted clause sets, respectively. Then, $\Phi = \Phi_s \cup \Phi_h$.

Let $vars(\Phi)$ be the set of all boolean variables in $\Phi$. A {\em truth assignment} is the interpretation of boolean variables into $0$ or $1$, i.e. $\mathcal{I} : vars(\Phi) \rightarrow \{0,1\}$. Then, the truth assignment for a set of weighted clauses $\Phi$ is:
\begin{equation}
    \mathcal{I}(\Phi) = \sum_{i=1}^m w_i (1 - \mathcal{I}(c_i)).
    \label{eq:wpms-interpretation}
\end{equation}

Therefore, WPMS is the problem of finding an {\em optimal assignment}  $\mathcal{I}^*(\Phi)$ \cite{ansotegui2013solving}, such that Equation~\ref{eq:wpms-interpretation} is minimized:
\begin{equation}
    \mathcal{I}^*(\Phi) = min\{ \mathcal{I}(\Phi) | \mathcal{I} : vars(\Phi) \rightarrow \{0,1\} \}.
    \label{eq:wpms-solution}
\end{equation}

The optimal assignment of Equation~\ref{eq:wpms-solution} could be inferred from $\mathcal{I}^*(vars(\Phi))$. Also, note that Equation~\ref{eq:wpms-interpretation} only iterates over soft clauses, since falsifying a hard clause would result in infinite cost and the solution would be invalid.

\subsection{\strips{} Action Model}
This work aims at recognizing the action models ${\cal A}$ in a domain represented with the \strips{} fragment of the Planning Domain Definition Language (PDDL) \cite{mcdermott1998pddl,haslum2019introduction} and negations. 
An {\it action schema} $\alpha \in {\cal A}$ is defined by a 3-tuple $\alpha=\langle \mathit{head}_\alpha, \mathit{pre}_\alpha, \mathit{eff}_\alpha  \rangle$ where:
\begin{itemize}
    \item $\mathit{head}_\alpha$ is the name of the action and a set of variables ${\cal V}' \subseteq {\cal V}$ referenced in $\mathit{pre}_\alpha$ and $\mathit{eff}_\alpha$,
    \item $\mathit{pre}_\alpha$ is a {\em well-formed formula} with a conjunction of literals that represent the action preconditions, 
    \item $\mathit{eff}_\alpha$ is the list of effects which consists of literals that are updated either to true (add/positive effect) or false (delete/negative effect). Often, $\mathit{eff}_\alpha$ is splitted into $\mathit{add}_\alpha$ and $\mathit{del}_\alpha$ lists.
\end{itemize}

\section{Online Action Recognition}
The problem of {\em Action Recognition} is a trivial task under certain assumptions such as fully observable transitions, finite set of relations and objects that are known to be true ({\em closed-world} assumption), and a complete knowledge library of \strips{} action models to represent the possible interactions between an agent and the world. From an observer perspective, the action applied by an agent would be easily identified with a naive linear algorithm that loops over the set of grounded actions. Nevertheless, the problem becomes more complex when the knowledge library only contains {\em action signatures} (set of action symbols with their arity) and transitions are partially observable. Then, the action models must be learned \cite{amir2008learning,mourao2012learning,aineto2018learning}. The problem is even harder when the knowledge library starts empty and the interpretation of the each relation between objects is unknown beforehand ({\em open-world} assumption) \cite{minker1982indefinite}. We focus on the online version of the latter paradigm. 

\subsection{Input: set of observations}

Inputs consist of observations over a set of transitions. Implicitly, each transition is a $ (s,a,s') $ triplet, where that $s$ is the previous state, $a$ is the action applied by an agent, and $s'$ is the successor state. However, in OAR, the action applied by the agent is never observed, so each transition observation is $o = (s, s')$. When the input consists of a set of observations $\mathcal{O}$, each observation $o \in \mathcal{O}$ is sequentially processed following the order in which they were observed. 


We use a {\em factored state} representation where each state $s$ is defined by a set of  predicates $p_n(d_1,\ldots,d_n)$ (or their negation), such that $p_n$ is a predicate symbol of arity $n$, and every $d_i  \in \mathcal{D}$ for $1 \leq i \leq n$ is an object in the universe.

In an {\em open world}, a grounded predicate $p_n(d_1,\ldots,d_n)$ is known to be true if each $d_i\in\mathcal{C}$ and it is explicitly true in a given state $s$ (resp. $\neg p_n(d_1,\ldots,d_n)$). Therefore, a state $s$ is {\em fully observable} if for every predicate and constants, there is an interpretation $\mathcal{I}_s(p_n(d_1,\ldots,d_n))$ which maps the grounded predicate either to true or false. Thus, a state $s$ is {\em partially observable} if there is no such interpretation $\mathcal{I}_s$ for all known predicates and constants.

Next, we introduce some definitions over $ \varphi = p_n(d_1,\ldots,d_n) $ that will become handy:
\begin{itemize}
    \item A function $ \typefn{\varphi} \defeq p_n $.
    \item A function $ \argfn{i}{\varphi} \defeq d_i $.
    \item A function $ \iota_s(\varphi) \defeq \mathrm{true} $ iff $ \mathcal{I}_s(\varphi) $ is defined, else $ \mathrm{false} $.
\end{itemize}

\subsection{Output: first-order {\strips} action model} 
%

The output for each observation $o \in \mathcal{O}$ is a grounded action $a$ composed of the action model $\alpha \in \mathcal{A}$ and its substitution $\sigma\in \Sigma$. We denote as $ {\cal D}_{\alpha} $ the set of objects (either constants or variables) referenced within $ \alpha $'s effects and preconditions. An action $a$ is said to be grounded or instantiated when the variables in $\mathit{head}_\alpha$ are substituted by a tuple of constants from $\mathcal{D}$ of equal size. We denote this by $ \alpha \sigma $, where $ \sigma \in \Sigma $ is an object substitution (and thus $ \Sigma $ is the set of all possible substitutions), as those introduced earlier. This nomenclature extends naturally to the substitution of arbitrary objects, not just variables. We adopt the convention that if $ \sigma $ does not define an explicit substitution for an object $ d \in {\cal D}_\alpha $, then its application leaves all references to $ d $ unchanged.


A {\bf solution} to the problem is then a {\em grounded action} $a$, such that $a=\alpha\sigma$ where $\alpha \in {\cal A}$ and $\sigma\in\Sigma$, which completes the observation $(s,s')$ into $(s,a,s')$ and the following conditions hold: 

\begin{condition}[Valid Preconditions]
The interpretation evaluates the preconditions to true, i.e. $\mathcal{I}_s(pre_{a}) = \top$, in other words, the action preconditions hold in the previous state, i.e. $pre_{a} \subseteq s$.
\label{cond:valid-prec}
\end{condition}

\begin{condition}[Valid Transition]
The successor state $s'$ is a direct consequence from applying $a$ in $s$, such that $s' = (s\setminus del_a)\cup add_a$.
\label{cond:valid-trans}
\end{condition}

\subsection{Problem statement} 
The OAR problem is tabular in that background theories, such as schemata ${\cal A}$, are unknown. Indeed, the problem is twofold, where the action is recognized if exists in the knowledge library, otherwise learns a new action model \cite{arora2018review} and unifies it with previous knowledge.

\begin{definition}[Online Action Recognition]
The OAR problem consists in finding a function $\mathcal{P} : \mathcal{O} \rightarrow \mathcal{A} \times \Sigma$ that sequentially maps each observation $o\in\mathcal{O}$ into an action $\alpha\in{\cal A}$, and a substitution $\sigma\in\Sigma$ s.t. Conditions~\ref{cond:valid-prec} and \ref{cond:valid-trans} hold. If $\alpha \notin {\cal A}$ initially, a new $\alpha$ must be learned. 
\label{def:oar}
\end{definition}


\section{Methodology}

The main algorithm is \thealgorithmacro{}. It starts with the set of predicate symbols but neither universe nor action library knowledge are known (i.e. ${\cal D} = \emptyset$ and ${\cal A} = \emptyset$). Both are greedily learned with every new observation $o=(s,s') $ as follows:
\begin{enumerate}
    \item For each pair of consecutive states $s$ and $s'$, it generates a Trivial Grounded Action (TGA) $a$ that explains this transition, but does not generalize.
    \item \thealgorithmacro{} merges $ a $ with the {\em closest} action in $\alpha \in \mathcal{A} $ via Action Unification (AU) using a weighted partial MaxSAT, which results in a new action $\alpha'$ and a grounding $\sigma$.
    \item \thealgorithmacro{} recognizes grounded action $ a' = \alpha'\sigma $ as the underlying action that produced observation $ o $.
\end{enumerate}


\subsection{Construction of Trivial Grounded Actions}


A TGA is a grounded \strips{} action $ a $  that explains just one transition from $ s $ to $ s' $. Hence, $ s' $ can be reached if $ a $ is applied to $ s $, but does not generalized to other states. This may be also denoted by $ s \xrightarrow{a} s'$.
To construct $a$ in a full observability setting, we set the conjunction of the predicates in $ s $ as the precondition, i.e. $\mathit{pre}_a = s$. The effects are defined as an add list $\mathit{add}_a = (s' \setminus s)$, and a delete list $\mathit{del}_a = (s \setminus s')$. To extend this basic construction to support partial observability, we consider uncertain predicates in $ s $ and $ s' $ as potentially true and false, and mark their occurrences within $\mathit{pre}_a$, $\mathit{add}_a$ and $\mathit{del}_a$ and effect as uncertain.

The preconditions and effects of any action can be joined into a single set $\setstyle{L}_a$ of \textit{labeled predicates}, each defined by a triplet $ \chi = (\varphi, l, k) $, where: 
\begin{itemize}
    \item $ \varphi = p_n(d_1,\ldots,d_n) $ is a predicate as defined earlier,
    \item $ l \in \{ \textsc{Pre}, \textsc{Add}, \textsc{Del} \} $ is the label to denote either the precondition, add or delete list,
    \item $ k $ is either true if the predicate is known to belong to $ l $ with total certainty, or false if it is unknown. 
\end{itemize}

Notice that $ \setstyle{L}_a $ is sufficient to represent $ a $, since the parameters of $ a $ can be constructed as the union of all the variables appearing in $ \setstyle{L}_a $ (i.e. all variables in $ {\cal D}_a $).

The $ k $ flag of a labeled predicate is set depending on whether $ \varphi $ is known in $ s $ and $ s' $.

Let us delve deeper in the details of the construction. Let $ \setstyle{P}_s $ be the set of positive known propositions of $ s $, and $ \setstyle{U}_s $ the set of unknown propositions. That is:
\begin{equation}
\begin{split}
    \setstyle{P}_s \defeq & \{ \varphi = p_n(d_1,\ldots,d_n)~|~\mathcal{I}_s(\varphi) = \top \} \\
    \setstyle{U}_s \defeq & \{ \varphi = p_n(d_1,\ldots,d_n)~|~\iota_s(\varphi) = \bot \}
\end{split}
\end{equation}

Then, from $ s $ and $ s' $ we create the TGA $ a $ that connects them as follows:
\begin{equation}
\begin{split}
    \setstyle{Pre}_a = & \{ (\varphi, \textsc{Pre}, \top)~|~\varphi \in P_s \} \cup \\
                      & \{ (\varphi, \textsc{Pre}, \bot)~|~\varphi \in U_s \} \\
    \setstyle{Add}_a = & \{ (\varphi, \textsc{Add}, \top)~|~\varphi \in P_{s'} \setminus (P_{s} \cup U_{s}) \} \cup \\
                      & \{ (\varphi, \textsc{Add}, \bot)~|~\varphi \in (P_{s'} \cap U_{s}) \cup (U_{s'} \setminus P_{s}) \}\\
    \setstyle{Del}_a = & \{ (\varphi, \textsc{Del}, \top~|~\varphi \in P_{s} \setminus (P_{s'} \cup U_{s'}) \} \cup \\
                      & \{ (\varphi, \textsc{Del}, \bot)~|~\varphi \in (P_{s} \cap U_{s'}) \cup (U_{s} \setminus P_{s'}) \}\\
    \setstyle{L}_a   = & \setstyle{Pre}_a \cup \setstyle{Add}_a \cup \setstyle{Del}_a
\end{split}
\label{eq:tga}
\end{equation}

\begin{example}
Consider \textit{sokoban}, and a predicate $ \varphi = {\tt at}({\tt agent}, {\tt pos}\text{-}{\tt 1}\text{-}{\tt 1}) $. Suppose $ \mathcal{I}_s(\varphi) = \mathrm{true} $, but $ \iota_{s'}(\varphi) = \mathrm{false}$ (i.e. the location of the \textit{agent} is known to be \textit{(1,1)} in $ s $, but this is not known with certainty in $ s' $). Then, the created TGA contains a labeled predicate $ \chi = (\varphi, \textsc{Del}, \mathrm{false}) $ because it is unknown whether $ \varphi $ is removed in the transition.
\end{example}

\subsection{Action Unification with Weighted Partial MaxSAT}

Let us focus on Action Unification (AU), \thealgorithmacro{}'s mechanism to merge a TGA into its action library.

AU's goal is to find an action $ \alpha_u $ that generalizes $ \alpha_1 $ and $ \alpha_2 $, \textit{if it exists}. We say that $ \alpha_u $ generalizes $ \alpha_1 $ and $ \alpha_2 $ if there are two substitutions $ \sigma_1, \sigma_2 $ such that $ \alpha_i' = \alpha_u \sigma_i $, $ \mathit{eff}_{\alpha_i'} = \mathit{eff}_{\alpha_i} $ and $ \mathit{pre}_{\alpha_i'} \models \mathit{pre}_{\alpha_i} $ for $ i \in \{1,2\} $. Intuitively, $ \alpha_u $ \textit{preserves} the effects of $ \alpha_1 $ and $ \alpha_2 $, while its preconditions are relaxations of $ \mathit{pre}_{\alpha_1} $ and $ \mathit{pre}_{\alpha_2} $, lifting some objects in the process. First, we seek to preserve as many predicates in the precondition as possible. Second, among all the $ \alpha_u $ actions with maximal preconditions, we want one with the least number of new parameters. This relaxation/lifting mechanism makes $ \alpha_u $ applicable in a wider range of situations. So far, we have described AU as a dual-objective optimization problem.

To perform AU given $ \alpha_1$ and $\alpha_2 $, we encode as a WPMS the problem of finding an injective partial function $ \tau : {\cal D}_{\alpha_1} \rightarrow {\cal D}_{\alpha_2} $ such that $ \tau(o_1) = o_2 $ implies that $ o_1 $ and $ o_2 $ will map to the same object in $ \alpha_u $. If $ o_1 = o_2 = o \in \mathcal{C} $, the reference to $ o $ is maintained within $ \alpha_u $ as a constant. Otherwise, $ o_1 $ and $ o_2 $ will be lifted. We say that two labeled predicates $ \chi_1 = (\varphi_1, l_1, k_1) \in \setstyle{L}_{\alpha_1} $ and $ \chi_2 = (\varphi_2, l_2, k_2) \in \setstyle{L}_{\alpha_2} $ \textit{match} iff $ l_1 = l_2 $, $ \typefn{\varphi_1} = \typefn{\varphi_2} = p_n $, and $ \tau(\argfn{i}{\varphi_1}) = \argfn{i}{\varphi_2} $ for $ 1 \leq i \leq n $. We denote as $\setstyle{M}_{\alpha_1,\alpha_2} $ the set of tuples $ (\chi_1, \chi_2) $ of potential matches. We say that a labeled predicate $ \chi \in \setstyle{L}_{\alpha_i} $ is \textit{preserved} if it has a match in the other action. A labeled predicate is preserved as certain if it is matched with a certain one. In order of priorities, our conditions for $ \tau $ are: (1) all certain effect predicate are preserved; (2) as many precondition and uncertain predicates as possible are preserved; and (3) as fewer parameters as possible are introduced.

In the WPMS encoding, denoted as $\Phi_{\alpha_1,\alpha_2}$, we use the following decision variables:
\begin{itemize}
    \item $x_{o_1,o_2}$ for $o_i\in{\cal D}_{\alpha_i}$, means that $\tau$ maps $o_1$ to $o_2$,
    \item $ y_{\chi_1,\chi_2}$ for $(\chi_1,\chi_2) \in \setstyle{M}_{\alpha_1,\alpha_2}$, means that $ \chi_1 $ matches $ \chi_2 $,
    \item $ z_{i,\chi_i}$ s.t. $i \in \{ 1,2 \}$, $\chi_i \in \setstyle{L}_{\alpha_i} $, means that $\chi_i$ is preserved.
\end{itemize}

We define four constraints that must be satisfied ({\bf H}ard), and two constraints to be optimized ({\bf S}oft):
\begin{itemize}
\item[] ({\bf H1}) $ \tau $ is an injective partial function, so $ \forall o_1, o_2 $:
\begin{equation}
\begin{split}
\text{At-Most-1}(\{x_{o_1,o_2'}~|~\forall o_2'\}), \\
\text{At-Most-1}(\{x_{o_1',o_2}~|~\forall o_1'\}).    
\end{split}
\end{equation}
    \item[] ({\bf H2}) Two potential matches $ (\varphi_1,\ldots) $ and $ (\varphi_2,\ldots) $, from $ \alpha_1 $ and $ \alpha_2 $ respectively, match iff $ \tau $ maps every $ i $th argument of $ \varphi_1 $ to the corresponding $ i $th argument of $ \varphi_2 $:
\begin{equation}
y_{\chi_1,\chi_2} \Leftrightarrow \bigwedge_i x_{\argfn{i}{\varphi_1},\argfn{i}{\varphi_2}}.
\end{equation}    
    \item[] ({\bf H3}) A labeled predicate $ \chi_i \in \setstyle{L}_{\alpha_i} $ is preserved iff it has at least one match in the other action:
\begin{equation}
\begin{split}
z_{1,\chi_1} \Leftrightarrow 
\quad \bigvee_{\chi_2'} y_{\chi_1,\chi_2'}, \\
\quad z_{2,\chi_2} \Leftrightarrow 
\quad \bigvee_{\chi_1'} y_{\chi_1',\chi_2}. 
\end{split}
\end{equation}    
    \item[] ({\bf H4}) Preserve non uncertain effects (i.e. $\chi_i = (\varphi_i, l_i, \mathrm{true}) \in \setstyle{L_{\alpha_i}}$, with $ l_i \neq \textsc{Pre} $):
\begin{equation}
z_{i,\chi_i}.
\end{equation}   
    \item[] \textbf{(S1)} (\textbf{Weight}=1) Avoid lifting objects, i.e. $ \forall o_1,o_2 \in \mathcal{C}$ s.t. $ o_1 \neq o_2 $:
\begin{equation}
\neg x_{o_1,o_2}.
\end{equation}
    \item[] \textbf{(S2)} (\textbf{Weight}=$ W_\mathit{big} $) Preserve preconditions and uncertain effects (i.e. $ \chi_i = (\varphi_i, l_i, k_i) $ with $ l_i = \textsc{Pre} \vee \neg k_i $):
\begin{equation}
z_{i,\chi_i}.
\end{equation}
\end{itemize}

For compactness, \textbf{(H2)} and \textbf{(H3)} have not been expressed in \textit{Clausal Normal Form} (CNF), but transforming them to CNF is trivial through the Tseitin transformation~\cite{tseitin1983complexity}. In \textbf{(H1)}, $ \text{At-Most-1} $ forbids more than one of the literals in the given set to become true. We use the quadratic encoding. Different weights are given to \textbf{(S1)} and \textbf{(S2)}. $ W_\mathit{big} $ must be large enough so that preserving predicates has priority over avoiding the introduction of parameters. $ W_\mathit{big} \geq \min(|{\cal D}_{\alpha_1}|, |{\cal D}_{\alpha_2}|) + 1 $ accomplishes this. Let us highlight that $ \mathcal{I}^*(\Phi_{\alpha_1,\alpha_2}) $ may be interpreted as a scaled distance between $ \alpha_1 $ and $ \alpha_2 $:
\begin{equation}
\label{eq:action-distance}
\begin{split}
    \mathcal{I}^*(\Phi_{\alpha_1,\alpha_2}) &= W_\mathit{big} \cdot N_\mathit{np} + N_\mathit{param}, \\
    \mathit{dist}_{\alpha_1,\alpha_2} &= \mathcal{I}^*(\Phi) / W_\mathit{big},
\end{split}
\end{equation}
where $ N_\mathit{np} $ is the number of non-preserved predicates, and $ N_\mathit{param} $ is the number of introduced parameters.
$ \mathit{dist}_{\alpha_1,\alpha_2} $ has an intuitive meaning: its integer part represents the number of eliminated predicates ($ N_\mathit{np} $) while its fractional part is proportional to the number of introduced parameters ($ N_\mathit{param} / W_\mathit{big}) $. If $ \alpha_1$ and $\alpha_2 $ cannot be unified, we define $ \mathit{dist}_{\alpha_1,\alpha_2} = \infty $.

\subsubsection{Complexity of Action Unification.}

We have proposed an approach for AU that requires a reduction to a WPMS problem. However, WPMS is known to be \textit{NP-Hard}, so it is reasonable to wonder if AU is in a more tractable complexity class. We claim that this is not the case.

\begin{thm}
Action Unification's problem is \textit{NP-Hard}.\footnote{Proof in appendix.}
\end{thm}

Despite the worst-case complexity of AU, we show that, in practice, real-time performance is possible.

\subsection{\thealgorithmfull}

\thealgorithmacro{} shows similarities to \textit{Hierarchical Clustering} (HC). Actions act as data points, and AU computes distances and builds clusters. Unlike in standard HC, we may cluster only actions whose effects can be preserved.

\begin{algorithm}[tb]
\caption{\thealgorithmacro{}}
\label{alg:action-recognition}
\hbox{\hspace*{\algorithmicindent}\textbf{Input:} Observation $ o = (s,s') $, action library $ \mathcal{A} $}
\hbox{\hspace*{\algorithmicindent}\textbf{Output:} Grounded action $ a' $ s.t. $ s \xrightarrow{a'} s'$}%
\begin{algorithmic}[1]
\STATE $ a \leftarrow \mathrm{BuildTGA}(s,s') $
\STATE $ \alpha \leftarrow \emptyset $, $ \alpha' \leftarrow \emptyset $, $ d_\mathit{min} \leftarrow \infty $
\FORALL{$ \beta \in \mathcal{A} $}
\STATE $ (\alpha_u, \mathit{dist}_{\beta,a}) \leftarrow \mathrm{ActionUnification}(\beta,a) $
\IF{$ \mathit{dist}_{\beta,a} < d_\mathit{min} $}
\STATE $ \alpha \leftarrow \beta $, $ \alpha' \leftarrow \alpha_u $, $ d_\mathit{min} \leftarrow \mathit{dist}_{\beta,a} $
\ENDIF
\ENDFOR
\IF{$ \alpha' \neq \emptyset $}
\STATE Remove $ \alpha $ from $ \mathcal{A} $
\STATE $ a' \leftarrow \alpha'\sigma $, with $ \sigma $ s.t. $ \mathit{eff}_{\alpha' \sigma} = \mathit{eff}_a $
\STATE Add $ \alpha' $ to $ \mathcal{A} $
\ELSE
\STATE $ a' \leftarrow a $
\STATE Add $ a $ to $ \mathcal{A} $
\ENDIF
\RETURN $ a' $
\end{algorithmic}
\end{algorithm}

\thealgorithmacro{}'s recognition subroutine is outlined in Algorithm~\ref{alg:action-recognition}. This subroutine updates $ |\mathcal{A}| $ on the basis of a new observation $ o = (s,  s') $, and explains $ s \rightarrow s' $ with a grounded action. It starts building a TGA $ a $ from $ o $ (line 1). After initializing some bookkeeping variables (line 2), it obtains, if possible, the closest action to $ a $ from $ |\mathcal{A}| $, and the result of AU (lines 3-8). If $ a $ could be unified to at least some $ \alpha \in \mathcal{A} $, $ \alpha $ is replaced by the unified action $ \alpha' $, and $ a' $ (the return value) is set to the grounding of $ \alpha' $ that fills the gap in $ o $ (lines 9-12). Otherwise, the TGA is assigned to the return value and added as is to $ |\mathcal{A}| $ (lines 13-16).

In practice, we also have a top-level procedure that initializes $\mathcal{A}$ to an empty set and runs Algorithm~\ref{alg:action-recognition} for each observation it encounters. \thealgorithmacro{} fills $\mathcal{A}$ progressively, and outputs the grounded action that explains each transition.

\section{Evaluation}

We have implemented\footnote{\url{https://github.com/sprkrd/sat_strips_learn}} and evaluated\footnote{Machine specifications: AMD Ryzen 7 3700X @ 3.6GHz, 32GB of DDR4 RAM CL15 @ 3200MHz.} \thealgorithmacro{} in a benchmark of 9 domains~\cite{muise2016planning.domains, Silver2020PDDLGym:Problems}. 
We have conducted two sets of experiments: with full and with partial observability. We use goal-oriented traces, i.e. list of pairs of consecutive states where a goal condition holds in the last state, so that all relevant actions are produced. 

\begin{table}[tb]
\caption{Results with (a) full and (b) partial observability. \textbf{T} is in milliseconds, \textbf{M} in MB, and \textbf{Prec.} and \textbf{Rec.} in \%.}
\label{tab:quantitative-results}
\begin{subtable}{\columnwidth}
\centering
\resizebox{\linewidth}{!}{%
\begin{tabular}{rrrrrrr}
\toprule
\textbf{Domain} & $\boldsymbol{|\mathcal{A}|}$ & $\boldsymbol{|\mathcal{O}|}$ &      \textbf{T} &             \textbf{M} &      \textbf{Prec.} &    \textbf{Rec.} \\ \midrule
         blocks &                            4 &                           96 &      $14 \pm 5$ &                    1.0 &         $100 \pm 1$ &       $100 \pm 0$\\
          depot &                            5 &                          300 &     $70 \pm 57$ &                    1.7 &         $92 \pm 12$ &        $96 \pm 5$\\
       elevator &                            3 &                          147 &     $31 \pm 32$ &                    1.6 &         $87 \pm 11$ &       $73 \pm 10$\\
        gripper &                            3 &                          262 &     $46 \pm 30$ &                    1.7 &         $100 \pm 3$ &       $100 \pm 0$\\
      minecraft &                            4 &                           23 &       $7 \pm 4$ &                    1.5 &          $97 \pm 6$ &       $100 \pm 0$\\
onearmedgripper &                            3 &                          284 &     $38 \pm 21$ &                    1.7 &         $100 \pm 3$ &       $100 \pm 0$\\
  rearrangement &                            4 &                           42 &     $19 \pm 11$ &                    1.7 &          $93 \pm 9$ &        $97 \pm 5$\\
        sokoban &                            4 &                          598 &     $49 \pm 24$ &                    1.7 &          $90 \pm 3$ &        $91 \pm 2$\\
         travel &                            5 &                           48 &     $22 \pm 17$ &                    1.9 &         $84 \pm 15$ &       $89 \pm 11$\\
\bottomrule
\end{tabular}}
\caption{}
\label{tab:quantitative-results-a}
\end{subtable}

\begin{subtable}{\columnwidth}
\centering
\resizebox{\linewidth}{!}{%
\begin{tabular}{rrrrrrr}
\toprule
\textbf{Domain} & $\boldsymbol{|\mathcal{A}|}$ & $\boldsymbol{|\mathcal{O}|} $ &     \textbf{T} &      \textbf{M} &      \textbf{Prec.} &    \textbf{Rec.} \\ \midrule
         blocks &                            4 &                            96 &     $16 \pm 7$ &             1.3 &         $90 \pm 15$ &        $99 \pm 6$\\
          depot &                            5 &                           300 &   $120 \pm 66$ &             3.8 &         $88 \pm 15$ &        $95 \pm 7$\\
       elevator &                            3 &                           147 &    $48 \pm 26$ &             5.3 &         $83 \pm 22$ &       $66 \pm 19$\\
        gripper &                            3 &                           262 &    $45 \pm 25$ &             5.2 &         $96 \pm 10$ &       $100 \pm 2$\\
      minecraft &                            4 &                            23 &    $30 \pm 27$ &             5.4 &         $65 \pm 23$ &        $99 \pm 6$\\
onearmedgripper &                            3 &                           284 &    $32 \pm 14$ &             5.4 &         $95 \pm 10$ &       $100 \pm 3$\\
  rearrangement &                            4 &                            42 &    $28 \pm 15$ &             5.5 &         $80 \pm 20$ &        $96 \pm 7$\\
        sokoban &                            4 &                           598 &  $158 \pm 146$ &              13 &         $89 \pm 12$ &        $86 \pm 5$\\
         travel &                            5 &                            48 &   $91 \pm 114$ &              16 &         $68 \pm 27$ &       $85 \pm 10$\\
\bottomrule
\end{tabular}}
\caption{}
\label{tab:quantitative-results-b}
\end{subtable}

\end{table}

For each domain, \thealgorithmacro{} observes the transitions that arise while solving 8 problems in succession. Each problem depicts a different number of objects and, thus, has a different impact in the number of variables and clauses that AU needs to encode\footnote{See appendix for domain and problem characteristics}. For each domain, we report:
\begin{itemize}
    \item $ \boldsymbol{|\mathcal{A}|} $: final size of \thealgorithmacro{}'s library.
    \item $ \boldsymbol{|\mathcal{O}|} $: total number of transitions across the 8 problems.
    \item \textbf{T}: Average CPU time taken by Algorithm~\ref{alg:action-recognition}.
    \item \textbf{M}: Peak memory usage for solving AU.
    \item \textbf{Prec.}: Precision of a recognized action $ a_g $, compared to the grounded action $ a_\textit{ref} $ that was used to perform the transition. It is the ratio of correct labeled predicates in $ a_g $:
\begin{equation}
    \textit{Prec}_{a_{ref}}(a_g) = 100 \frac{|\setstyle{L}_{a_g} \cap \setstyle{L}_{a_\textit{ref}}|}{|\setstyle{L}_{a_g}|}.
\end{equation}
    \item \textbf{Rec.}: Average recall (\%) of $ a_g $ compared to $ a_\textit{ref} $. It is the ratio of predicates in $ a_\textit{ref} $ that have been captured by $ a_g $:
\begin{equation}
    \textit{Rec}_{a_\textit{ref}}(a_g) = 100 \frac{|\setstyle{L}_{a_g} \cap \setstyle{L}_{a_\textit{ref}}|}{|\setstyle{L}_{a_\textit{ref}}|}.
\end{equation}
\end{itemize}
\textbf{T} and \textbf{M} are performance-related (quantitative evaluation). \textbf{Prec.} and \textbf{Rec.} measure the correctness and completeness of $ a_g $ (qualitative evaluation). We report both evaluations for full and partial observability in Table~\ref{tab:quantitative-results}.

\begin{figure}[tb]
    \begin{subfigure}{0.49\columnwidth}
    \includegraphics[width=\linewidth]{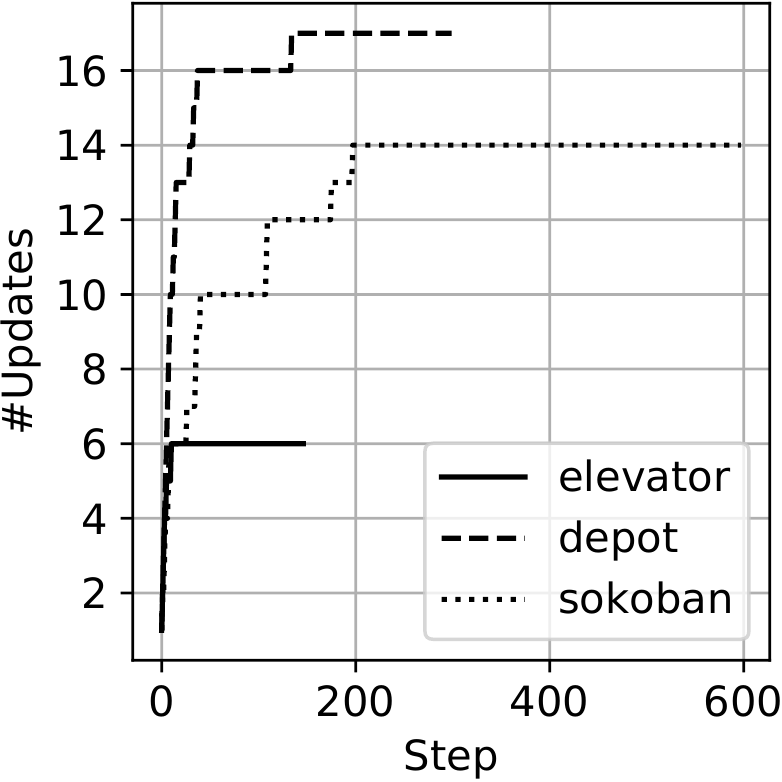}
    \caption{}
    \label{fig:update-accumulation-a}
    \end{subfigure}
    \hfill
    \begin{subfigure}{0.49\columnwidth}
    \includegraphics[width=\linewidth]{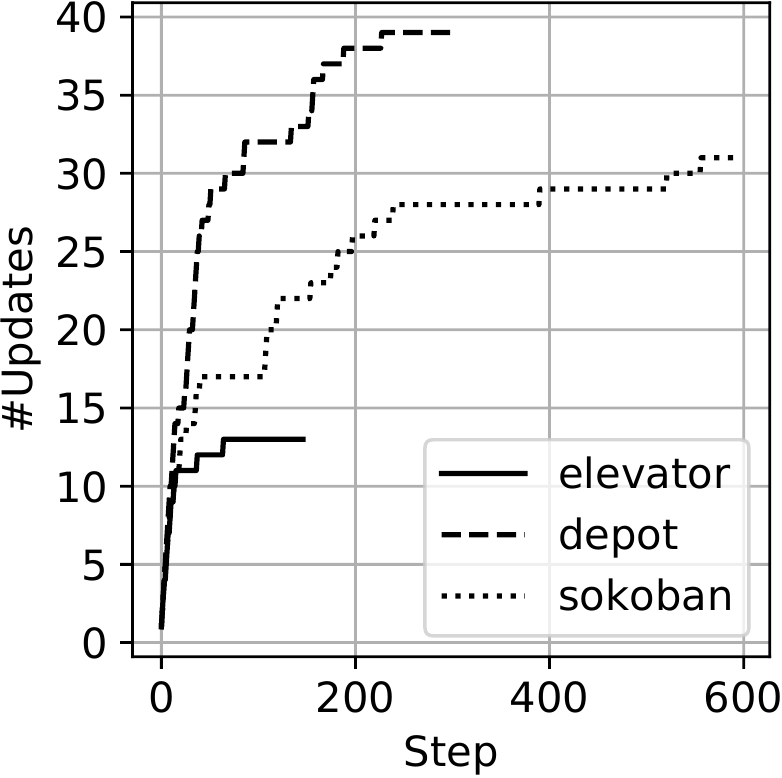}
    \caption{}
    \label{fig:update-accumulation-b}
    \end{subfigure}
    \caption{Accumulated updates for three domains in (a) full and (b) partial observability.}
    \label{fig:update-accumulation}
\end{figure}

Let us focus first on the full-observability results (Table~\ref{tab:quantitative-results-a}). Times are in general well below 1 second. Therefore, our claim about the real-time potential of \thealgorithmacro{} holds, as long as the throughput of observations is reasonably limited (e.g. a few seconds between observations). Memory requirements are also within a reasonable bound, requiring in the order of a few MBs to solve AU. However, we have found that some domains are challenging to \thealgorithmacro{} when it has just started with an empty $ \mathcal{A} $. For instance, \thealgorithmacro{} cannot cope with \textit{gripper} in a timely fashion if it starts with a problem with 20 objects or more. This is understandable, because the size of the WPMS encoding grows quadratically with the matches of predicates and objects. We believe this is not a huge setback, since it is very natural to ramp up the number of objects progressively. Additional performance is achieved through some optimizations not described here, like a \textit{broad phase} stage in Algorithm~\ref{alg:action-recognition} to avoid a full call to AU when it is obvious that actions cannot be unified. 

We see in general that the recall is very large. The reason lies in how \thealgorithmacro{} works. In deterministic settings, effects are almost always identified with high accuracy, while preconditions are relaxed as little as possible. In 5 cases the recall is lower than 100\%. In 3 of those, moreover, $|\mathcal{A}|$ is not equal to the number of actions in the expert's domain or \textit{Ground Truth Model} (GTM). In \textit{elevator} (GTM with 4 actions), there are two very similar actions for going up and down that only differ in one predicate of the precondition. \thealgorithmacro{} recognizes them as a generic move up or down action and discards that predicate. On the other hand, \textit{sokoban} (GTM with 3 actions) contains an action for moving stones to non-goal positions that deletes an $ \textit{at-goal}(\cdot) $ predicate which is not necessarily in the state. Thus, its deletion is not always observed, and \thealgorithmacro{} fills $ |\mathcal{A}| $ with two different actions that address different contexts. \textit{Travel} (GTM with 4 actions) presents a similar situation, but for adding an already present predicate. The 2 remaining cases (\textit{rearrangement} and \textit{depot}) have lower than 100\% recalls for similar reasons, but their $ |\mathcal{A}| $ is equal to their GTM's.

The precision is not as high as the recall because \thealgorithmacro{} holds onto precondition predicates that are not present in the GTM, but are confirmed by all the observed transitions. For instance, grid-based worlds with adjacency predicates, like \textit{sokoban}, show symmetries. To move from \texttt{pos-5-6} to \texttt{pos-6-6}, a predicate such as \texttt{move-dir}(\texttt{pos-5-6}, \texttt{pos-6-6}, \texttt{dir-right}) must hold true. However, \thealgorithmacro{} observes that, whenever that happens, the predicate \texttt{move-dir}(\texttt{pos-6-6}, \texttt{pos-5-6}, \texttt{dir-left}) is also true. However, the GTM lists only one adjacency precondition. In most cases, \thealgorithmacro{} is not incorrect from a semantic standpoint, but the precision metric counts it as an error.

Table~\ref{tab:quantitative-results-b} shows the results for partial observability. These have been obtained making parts of the state ignored by \thealgorithmacro{} at random. Namely, the interpretation of from 0 to 5 predicates has been removed. Results are averaged over 5 executions of the 8 problems for all the domains. We see penalties in the performance. This is most evident in \textit{sokoban}, because all problems present more than 50 objects. The introduction of uncertain predicates drives up the number of potential matches significantly. Other domains show smaller increases in recognition times and memory requirements. Precision and recall have suffered, but are not significantly worse than the full observability results. We have also experimented with greater proportions of unknown predicates (0 to 10). \thealgorithmacro{} starts requiring large computational times for AU in \textit{elevator} and \textit{sokoban} (in the order of several minutes). As we would expect, partial observability has a much larger computational burden. A way to address all these problems is to start with smaller problems and build $ |\mathcal{A}| $ progressively.

Figure~\ref{fig:update-accumulation} shows how often $ \mathcal{A} $ is updated (via action addition or replacement) in three domains. The X axis shows the number of steps taken, while the Y axis shows the accumulated number of updates. We consider that an update is made only when a TGA is not entirely subsumed by an action already present in $ \mathcal{A} $ (i.e. either a parameter is added or a labeled predicate is removed). Notice that $ \mathcal{A} $ stabilizes early on for \textit{elevator}. For \textit{sokoban}, it plateaus half-way towards the final number of steps, and \textit{depot} at approximately two thirds. The drop in the rise rate of the curves shows that, early on, $ \mathcal{A} $ is empty and is updated rather frequently with new observations. However, as $ \mathcal{A} $ is filled, \thealgorithmacro{} sees more actions and comes up with good general schemata that generalize all the transitions that could happen. The shapes of the full and partial observability curves are very similar. The most notable differences are in the scale and that the partial observability curves do not entirely plateau, but still exhibit some small jumps towards the final steps. This is intuitive: partial observations cause an increased number of updates, and that $ \mathcal{A} $ does not stabilize until later.

\section{Discussion and Conclusions}
In this paper, we have proposed \thealgorithmacro{}, an algorithm for recognizing {\strips} action models from partially observable state transitions. It uses AU to merge observations into its action library, constructing an action hierarchy and improving its action models through generalization. \thealgorithmacro{} shows a high computing performance and the recognized actions have strong similarities to expert handcrafted models. Thus, \thealgorithmacro{} ability to learn without action signatures makes it a promising contender to other model learning approaches.

As future work we consider to learn more expressive actions with generalized planning \cite{jimenez2019review}, and refine actions with negative examples \cite{segovia2020generalized}. Allow noisy observations extracted from sensor sampling~\cite{yang2009activity} which would make it suitable for more realistic applications on the wild. Also recognizing hidden variables and intermediate states in action sequences with Bayesian inference~\cite{aineto2020observation}. Finally, we think there is a great potential for an application-ready methodology adopting \thealgorithmacro{}'s philosophy to learn from low-level data, i.e. robot motions~\cite{konidaris2018skills}, images~\cite{asai2018classical}  or graphs~\cite{bonet2019learning}. 

\section{ Acknowledgments}
The research leading to these results has received funding from the EU H2020 research and innovation programme under grant agreement no.731761, IMAGINE. Javier Segovia-Aguas was also partially supported by TAILOR, a project funded by EU H2020 research and innovation programme no. 952215, an ERC Advanced Grant no. 885107, and grant TIN-2015-67959-P from MINECO, Spain.

\bibliography{bib-aaai21}

\clearpage

\appendix
\appendixpage

\section{Action Unification Problem is NP-Hard}

In the paper we present Action Unification (AU) both as a problem and an algorithm. In this section, we refer to the former, proving {\bf Action Unification problem} is {\bf NP-Hard} as stated in Theorem 1. The proof uses a polynomial-time reduction from 3-SAT, which is an NP-Complete problem \cite{karp1972reducibility}, to AU.

\begin{definition}

3-SAT is a decision problem over a set of propositional variables $\mathrm{V}$ (i.e. $\{x_1,\ldots,x_n\}$) of a Boolean formula, which is expressed as a conjunction of clauses $\mathrm{C}$. Then, each clause $c\in \mathrm{C}$ is a disjunction of exactly 3 literals, i.e. $c=(l_{c,1}\vee l_{c,2} \vee l_{c,3} )$, where a literal $l$ is either a variable $x\in\mathrm{V}$ or its negation $\neg x$.
The output of 3-SAT is {\bf yes} if at least one truth assignment of variables $\mathrm{V}$ exists such that every $c \in \mathrm{C} $ evaluates to true, otherwise is \textbf{no}.
\end{definition}

\begin{proof}

3-SAT inputs are defined as propositional variables $\mathrm{V}$ and clauses $\mathrm{C}$, while in AU they consist of two \textsc{Strips}-like actions. Next, we show how to polynomically reduce 3-SAT to AU, so that solving AU also solves the former:

\begin{enumerate}
    \item Let us denote as $ \mathrm{var}(l) $ the propositional variable related to $l$ (i.e. $ \mathrm{var}(x) = \mathrm{var}(\neg x) = x$). Then, we create an $n$-ary action $ a_1(x_1,\ldots,x_n) $ whose parameters are all variables $ x \in \mathrm{V} $. For each clause $ c = ( l_{c,1} \vee l_{c.2} \vee l_{c,3} ) \in \mathrm{C} $, we include in $\alpha_1$'s precondition a ternary predicate $ c_3(\mathrm{var}(l_{c,1}), \allowbreak \mathrm{var}(l_{c,2}), \allowbreak \mathrm{var}(l_{c,3})) $. This predicate means that variables listed within its arguments are in $ c $. However, it does not state whether they appear in positive or negative form. For instance, for $ c = (x_1 \vee \neg x_2 \vee \neg x_3) $, we create a precondition predicate $ c_3 ( x_1, x_2, x_3 ) $, because $ x_1 $, $ x_2 $ and $ x_3 $ are all the variables within $ c $. The add and delete lists of $ \alpha_1 $ are empty. 
    \item We create a second action $ \alpha_2 $ without parameters. For each clause $ c = (l_{c,1} \vee l_{c.2} \vee l_{c,3}) \in \mathrm{C} $, and for all the truth assignments 
    $ \mathrm{ta}_{\mathrm{var}(l_{c,i})} \in \{ \mathrm{true}_{\mathrm{var}(l_{c,i})}, \allowbreak \mathrm{false}_{\mathrm{var}(l_{c,i})} \} $ to variables $ \mathrm{var}(l_{c,i}) \in c $ that make $ c $ true, we include a precondition $ c_3(\mathrm{ta}_{\mathrm{var}(l_{c,1}}),\mathrm{ta}_{\mathrm{var}(l_{c,2})},\mathrm{ta}_{\mathrm{var}(l_{c,3})}) $. For instance, given $ c = (x_1 \vee \neg x_2 \vee \neg x_3) $ we would include the following 7 preconditions in $a_2$:
    \begin{itemize}
        \item $ c_3(\mathrm{false}_{x_1},\mathrm{false}_{x_2},\mathrm{false}_{x_3}) $
        \item $ c_3(\mathrm{false}_{x_1},\mathrm{false}_{x_2},\mathrm{true}_{x_3}) $
        \item $ c_3(\mathrm{false}_{x_1},\mathrm{true}_{x_2},\mathrm{false}_{x_3}) $
        \item $ c_3(\mathrm{true}_{x_1},\mathrm{false}_{x_2},\mathrm{false}_{x_3}) $
        \item $ c_3(\mathrm{true}_{x_1},\mathrm{false}_{x_2},\mathrm{true}_{x_3}) $
        \item $ c_3(\mathrm{true}_{x_1},\mathrm{true}_{x_2},\mathrm{false}_{x_3}) $
        \item $ c_3(\mathrm{true}_{x_1},\mathrm{true}_{x_2},\mathrm{true}_{x_3}) $
    \end{itemize}
    These preconditions list all the possible assignments to $ x_1 $, $ x_2 $ and $ x_3 $, except $ x_1 = \mathrm{false} $, $ x_2 = \mathrm{true} $, and  $ x_3 = \mathrm{true} $, which falsifies $ c $. Notice the inclusion of grounded terms (i.e. constants) $ \{ \mathrm{true}_{x}, \mathrm{false}_{x} \}$ for each $ x \in \mathrm{V} $.
\end{enumerate}

Performing AU between $ \alpha_1 $ and $ \alpha_2 $, we get an action $ \alpha_u $. The answer to the original 3-SAT problem can be derived checking if all the preconditions in $ a_1 $ have been preserved\footnote{A similar proof can be given to reduce 3-MaxSAT, the optimization variant of 3-SAT, to AU. The answer to 3-MaxSAT would directly be the number of preserved predicates.}, that is, whether $ | \mathit{pre}_{\alpha_1} | = | \mathit{pre}_{\alpha_u} | $. The rationale is that, in order to preserve the maximum number of preconditions, the lifted predicates in $ \alpha_1 $ have to be matched to the grounded ones in $ \alpha_2 $. This forces matching each variable $ x \in \mathrm{V} $ to an assignment $ \mathrm{ta}_{x} \in \{ \mathrm{true}_x, \mathrm{false}_x \} $, which has to be necessarily consistent among all clauses.

\textbf{The reduction is polynomial}. Action $ \alpha_1 $ is created with $ |\mathrm{V}| $ parameters and $ |\mathrm{C}| $ preconditions, each being a ternary predicate. Action $ \alpha_2 $ is created with $ 2 |\mathrm{V}| $ references to grounded terms, and with $ 7 | \mathrm{C} | $ ternary precondition predicates. Then, necessarily, \textbf{AU must be at least as hard as 3-SAT}, since a polynomial solution to AU would also solve 3-SAT in deterministic polynomial time. Since 3-SAT is known to be NP-Hard (more specifically, NP-Complete), then AU problem is also NP-Hard. Notice that AU is an optimization problem, not a decision one, so it cannot be in NP. Thus, AU problem is NP-Hard, but not NP-Complete.
\end{proof}

\begin{figure*}[tb!]
    \centering
    \includegraphics[width=0.9\linewidth]{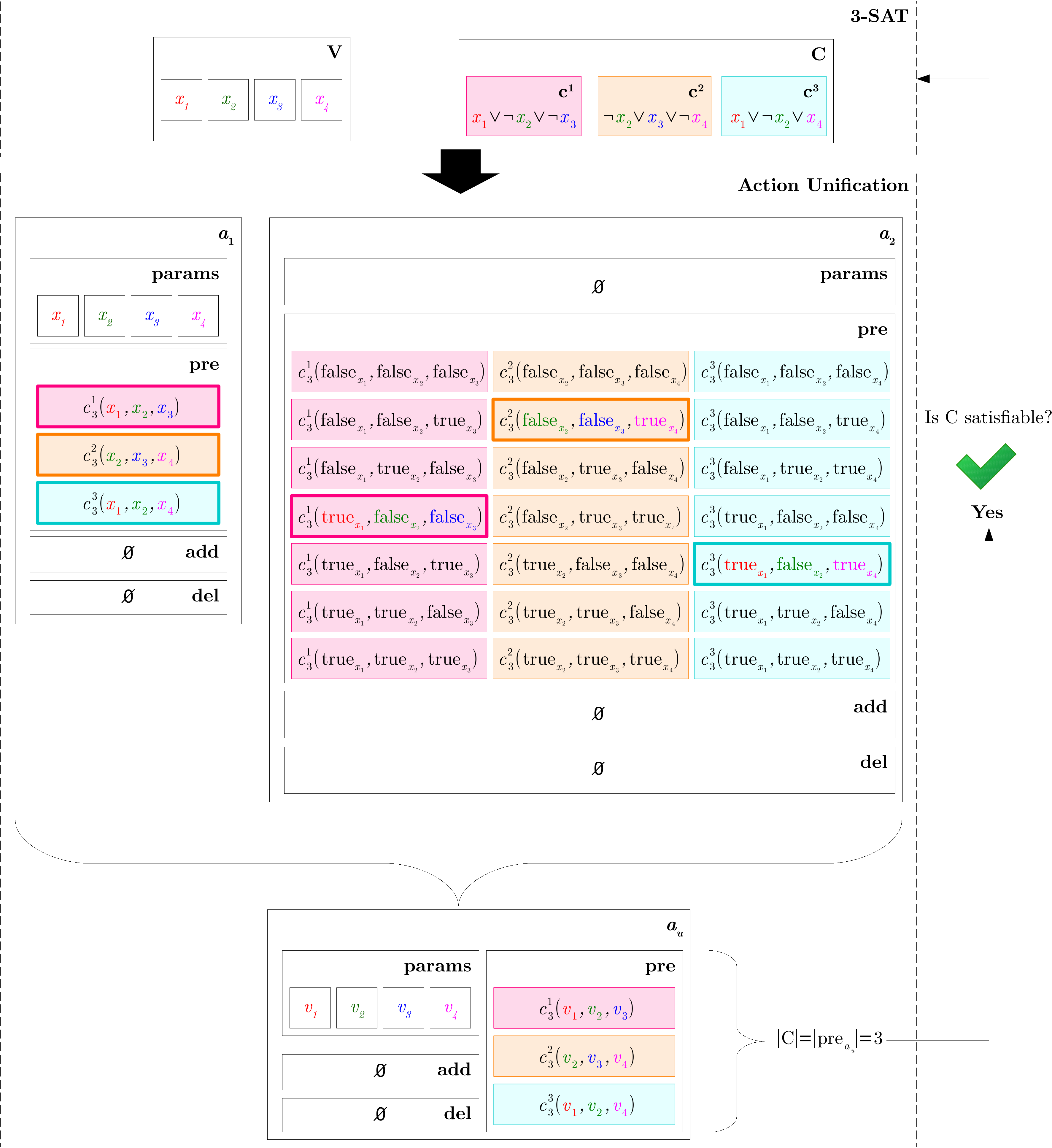}
    \caption{Polynomial-time reduction from a 3-SAT problem to an Action Unification problem.}
    \label{fig:3sat-to-au}
\end{figure*}

Let us provide an intuitive example to this proof.
\begin{example}
Let us consider a 3-SAT problem in which $ \mathrm{V} = \{ x_1, x_2, x_3, x_4 \} $ and $ \mathrm{C} = \{ (x_1 \vee \neg x_2 \vee \neg x_3), \allowbreak (\neg x_2 \vee x_3 \vee \neg x_4), \allowbreak (x_1 \vee \neg x_2 \vee x_4) \}$. The reduction from 3-SAT to AU is depicted in Figure~\ref{fig:3sat-to-au}. The original 3-SAT problem is at the top, while the reduction to AU is at the bottom. $ \alpha_1 $, depicted at the left, contains one precondition predicate per clause. Each one lists the variables that appear within the associated clause (predicates associated to the same clause have the same background color). At the right, $ \alpha_2 $ has, for each clause and each truth assignment that renders it true, a precondition that represents such assignment. To derive a solution to the original 3-SAT problem, we may check that all the predicates within $ \alpha_1 $ can be preserved with AU. There are as many ways to preserve the predicates as there are assignments that simultaneously satisfy all the clauses in $\mathrm{C}$. In the figure, we have represented one of such ways, highlighting the preserved predicates and colouring the truth assignments with the color of the associated variable. Thus, the represented assignment is $ x_1 = \mathrm{true} $, $ x_2 = \mathrm{false} $, $ x_3 = \mathrm{false} $, and $ x_4 = \mathrm{true} $.
\end{example}

\section{Distance vs Probability}
Computing the full probability distributions to recognize plans, activities and intents (PAIR) \cite{sukthankar2014plan} is very expensive, so methods in PAIR often approximate the recognition problem via either sampling or distance metrics. In our case, we are just interested in the most likely action, that we approximate as the action that is {\em closest} (unification distance) to the given observation (see Equation 9). In \citeauthor{aineto2019model} (\citeyear{aineto2019model}) is shown that computing the {\em closest consistent model} by distance minimization solves the same problem as the most likely model that explains the observation, similarly in {\em Online Action Recognition} the action which minimizes the unification distance is the same as the most likely action that explains the observation.

\section{Supplementary OARU Evaluations}

\begin{table}[tb!]
\centering
\caption{Domain characteristics}
\label{tab:domain-characteristics}
\resizebox{\columnwidth}{!}{%
\begin{tabular}{rrrr}
\toprule
         \textbf{Domain} & $\boldsymbol{|\mathcal{A}_{\mathit{GMT}}|}$ & \textbf{Max. Act. Arity} & \textbf{Max. Pred. Arity}\\ \midrule
         \textbf{blocks} &                                           4 &                        3 &                         2\\
          \textbf{depot} &                                           5 &                        4 &                         4\\
       \textbf{elevator} &                                           4 &                        2 &                         2\\
        \textbf{gripper} &                                           3 &                        3 &                         3\\
      \textbf{minecraft} &                                           5 &                        3 &                         2\\
\textbf{onearmedgripper} &                                           3 &                        3 &                         3\\
  \textbf{rearrangement} &                                           4 &                        4 &                         2\\
        \textbf{sokoban} &                                           3 &                        6 &                         3\\
         \textbf{travel} &                                           4 &                        4 &                         3\\
\bottomrule
\end{tabular}}
\end{table}

\begin{table}[tb!]
    \centering
    \caption{Number of objects per domain and per problem.}
    \label{tab:object-count}
\resizebox{\columnwidth}{!}{%
\begin{tabular}{rrrrrrrrr}
\toprule
         \textbf{Domain} & \textbf{ 1 } & \textbf{ 2 } & \textbf{ 3 } & \textbf{ 4 } & \textbf{ 5 } & \textbf{ 6 } & \textbf{ 7 } & \textbf{ 8 }\\ \midrule
         \textbf{blocks} &            5 &            6 &            6 &            7 &            7 &            7 &            5 &            6\\
          \textbf{depot} &           13 &           26 &           15 &           17 &           19 &           30 &           35 &           30\\
       \textbf{elevator} &            3 &            9 &           12 &           15 &           18 &           30 &           21 &           24\\
        \textbf{gripper} &            8 &           12 &           16 &           20 &           24 &           10 &           14 &           18\\
      \textbf{minecraft} &           23 &           27 &           25 &           29 &           23 &           23 &           25 &           22\\
\textbf{onearmedgripper} &            6 &            7 &            9 &           11 &           13 &           15 &           17 &           19\\
  \textbf{rearrangement} &           14 &           19 &           25 &           18 &           21 &           15 &           19 &           14\\
        \textbf{sokoban} &           56 &           80 &           57 &           86 &           99 &           79 &           49 &           77\\
         \textbf{travel} &           22 &           11 &           15 &           20 &           15 &            8 &           14 &           12\\
\bottomrule
\end{tabular}}
\end{table}

\begin{table*}[tb!]
    \centering
    \caption{Statistics about the state size encountered while solving the problems of each domain, measured as the number of active predicates (i.e. grounded predicates that evaluate to true). For each problem, the minimum/median/maximum size is reported.}
    \label{tab:state-size}
\resizebox{\linewidth}{!}{%
\begin{tabular}{rrrrrrrrrr}
\toprule
         \textbf{Domain} & \textbf{ 1 } & \textbf{ 2 } & \textbf{ 3 } & \textbf{ 4 } & \textbf{ 5 } & \textbf{ 6 } & \textbf{ 7 } & \textbf{ 8 } & \textbf{All}\\ \midrule
         \textbf{blocks} &     11/12/14 &     14/14/16 &     13/14/14 &     15/16/17 &     15/16/17 &     16/17/20 &     11/12/13 &     13/14/16 &     11/15/20\\
          \textbf{depot} &     58/60/62 &  122/124/128 &     68/71/74 &     81/83/86 &     92/95/98 &  146/148/152 &  171/175/182 &  144/148/152 &   58/147/182\\
       \textbf{elevator} &        7/8/8 &     31/33/34 &     49/52/53 &     71/76/76 &   97/101/103 &  241/249/251 &  127/133/134 &  161/169/169 &    7/132/251\\
        \textbf{gripper} &     13/14/15 &     21/22/23 &     29/30/31 &     37/38/39 &     45/46/47 &     17/18/19 &     25/26/27 &     33/34/35 &     13/34/47\\
      \textbf{minecraft} &     34/34/34 &     37/38/38 &     42/42/42 &     43/44/44 &     35/36/36 &     36/36/36 &     40/40/40 &     38/38/39 &     34/38/44\\
\textbf{onearmedgripper} &     10/11/11 &     12/13/13 &     16/17/17 &     20/21/21 &     24/25/25 &     28/29/29 &     32/33/33 &     36/37/37 &     10/28/37\\
  \textbf{rearrangement} &     19/19/19 &     28/28/28 &     36/36/36 &     23/23/23 &     32/32/32 &     22/22/22 &     28/28/28 &     19/19/19 &     19/28/36\\
        \textbf{sokoban} &  186/187/188 &  280/280/283 &  193/194/196 &  323/324/327 &  390/391/394 &  296/297/298 &  152/153/154 &  306/311/314 &  152/297/394\\
         \textbf{travel} &     66/67/68 &     31/33/34 &     44/45/47 &     51/51/52 &     46/46/47 &     23/24/25 &     43/45/46 &     29/29/29 &     23/45/68\\
\bottomrule
\end{tabular}}
\end{table*}

Table~\ref{tab:domain-characteristics} reports the characteristics of the 8 domains used for benchmarking. We report the size of the expert's action library, the maximum action arity and the maximum predicate arity. Table~\ref{tab:object-count} reports, for each domain, the number of objects that each of the considered problems features. Table~\ref{tab:state-size} reports, for each domain, the minimum, median and maximum sizes of the states encountered while solving each problem. All these metrics are indicative of the difficulty that AU has in those domains.

In our paper, have reported average quality and performance metrics for \textit{Online Action Recognition through Unification} (OARU) under full and partial observability when fed with goal-oriented observations that result from solving 8 problems. For partial observability, 5 runs have been averaged to account for the randomness from stripping away the interpretation of 0 to 5 predicates, sampled uniformly.

In general, we find that smaller problems (i.e. with low amount of objects and active predicates) are easy for AU, while the difficulty of AU in large problems is highly dependent on the structure of the domain. For instance, \textit{depot} and \textit{sokoban} have the largest computational times, both under full and under partial observability. Also, with partial observability, \textit{sokoban} has the second highest consumption of memory. On the other hand, \textit{elevator}'s computational load is quite low, even with its high object and predicate count. Contrarily to \textit{sokoban}, \textit{elevator} has actions and predicates with low arity. For domains with few objects and predicates, however, the computational times and memory consumption are quite low, with the exception of \textit{travel} under partial observability. This suggests that the conditions of partial observability are more harmful for some domains than others.

\subsection{Action Convergence and Recognition}

\begin{figure*}[tb!]
    \centering
    \includegraphics[width=\linewidth]{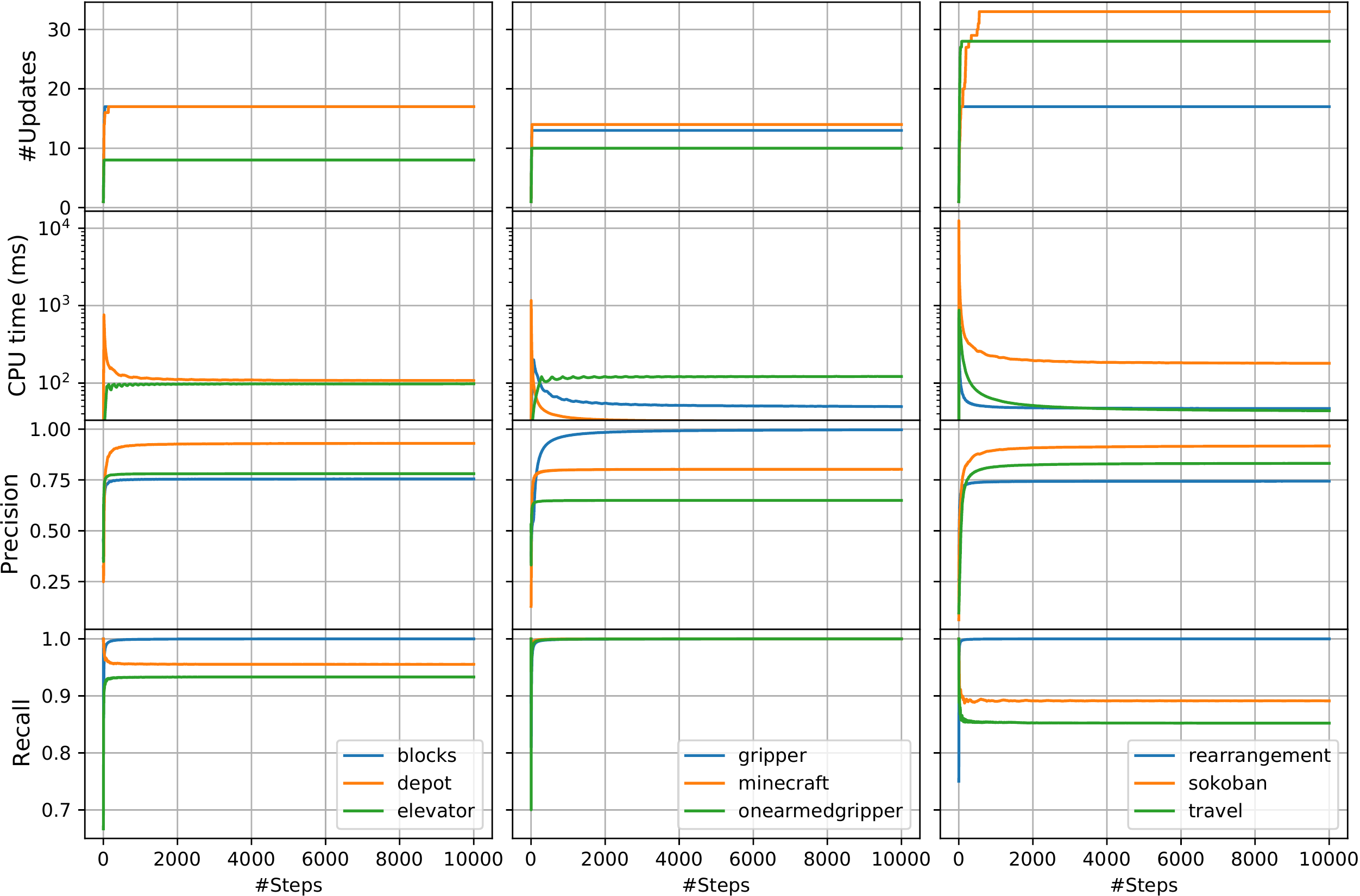}
    \caption{Metric evolution over time (partial observability)}
    \label{fig:action-convergence}
\end{figure*}

Figure~\ref{fig:action-convergence} shows the evolution of the number of updates, the CPU time and the precision and recall metrics over time, for all the 9 domains. This plot has been generated cycling through all the observations obtained after solving the 8 problems of all the domains, until reaching a fixed amount of steps (10000). We have chosen the partial observability setting described above, to address the most general type of problems that OARU may face.

Several interesting phenomena arise:
\begin{enumerate}
    \item The number of updates plateaus after a certain amount of observations. This is indicative of OARU's library stabilizing. Further observations do not update the library, and are completely subsumed by it. Even though under partial observability this happens later than under full observability, now OARU makes enough observations to achieve this stability.
    \item Computational times usually start high, but they decrease progressively as the initial \textit{Trivial Grounded Actions} (TGAs) are generalized through AU. AU effectively relaxes the preconditions, dispels the uncertainty of some predicates, and dismisses some other uncertain predicates. This means that there will be fewer potential matches in future unifications. Thus the number of variables and clauses in the WPMS encoding will also become smaller.
    \item The precision starts low, but then increases with the number of observations. This is natural, since observations contribute to relax overly restrictive preconditions and dispel spurious predicates. Even if the precision does not reach 100\%, the recognized actions are not necessarily bad, as we analyzed in the discussion of our paper.
    \item Conversely, the recall starts high, because the initial TGAs are very specific. This has the advantage that the TGAs do not miss the predicates present in the expert's domain. However, as they are generalized, some predicates may be lost. This can be justified in a per-case basis, as we have done in the paper's discussion. However, in this particular run, it does not fall below 80\%.
\end{enumerate}


\subsection{Learned Action Libraries}
In this paper, actions are recognized from a learned action library. These are the final action libraries for each domain from a random execution with partial observability\footnote{Original domains are in \url{https://github.com/tomsilver/pddlgym}.}. We have interpreted the name of actions and parameters, to ease their understanding. Uncertain predicates are represented with an appended question mark ``?'', and parameters with the form of zN (N is a possibly empty number) are used either in redundant or static preconditions that have not been removed because such event has never been observed in a transition.

\FloatBarrier

\subsubsection{Blocksworld.} OARU recognizes the original domain but with a redundant precondition in Unstack, Put-Down and Pick-Up. Parameters x and y stands for blocks while r is the robot gripper. Note that parameterizing actions with the robot, makes the domain general for multiple grippers.\\

\begin{tcolorbox}[breakable]
\begin{center}{\bf Blocksworld}\end{center}
\begin{flushleft}
\underline{Unstack(z, y, r, x):}\\
\texttt{Pre:} \nohyphens{block(z), block(y), block(x), clear(x), handempty(r), on(x,y), robot(r)}\\
\texttt{Add:} \nohyphens{clear(y), handfull(r), holding(x)}\\
\texttt{Del:} \nohyphens{clear(x), handempty(r), on(x,y)}\\
\end{flushleft}

\begin{flushleft}
\underline{Put-Down(z, r, x):}\\
\texttt{Pre:} \nohyphens{block(z), block(x), handfull(r), holding(x), robot(r)}\\
\texttt{Add:} \nohyphens{clear(x), handempty(r), ontable(x)}\\
\texttt{Del:} \nohyphens{handfull(r), holding(x)}\\
\end{flushleft}

\begin{flushleft}
\underline{Pick-Up(z, x, r):}\\
\texttt{Pre:} \nohyphens{block(z), block(x), clear(x), handempty(r), ontable(x), robot(r)}\\
\texttt{Add:} \nohyphens{handfull(r), holding(x)}\\
\texttt{Del:} \nohyphens{clear(x), handempty(r), ontable(x)}\\
\end{flushleft}

\begin{flushleft}
\underline{Stack(r, x, y):}\\
\texttt{Pre:} \nohyphens{block(x), block(y), clear(y), handfull(r), holding(x), robot(r)}\\
\texttt{Add:} \nohyphens{clear(x), handempty(r), on(x,y)}\\
\texttt{Del:} \nohyphens{clear(y), handfull(r), holding(x)}\\
\end{flushleft}

\end{tcolorbox}

\subsubsection{Depot.} The 5 action schemas from the original domain are recognized, with some redundacies in the locations of objects. Variables p, from and to represent places, c is a crate, h a hoist, s a surface and t a truck.

\begin{tcolorbox}[breakable]
\begin{center}{\bf Depot}\end{center}
\begin{flushleft}
\underline{Lift(p, z, c, h, s):}\\
\texttt{Pre:} \nohyphens{at(c,p), at(h,p), at(s,p), available(h), clear(c), crate(c), hoist(h), locatable(c), locatable(h), locatable(s), object(p), object(z), object(c), object(h), object(s), on(c,s), place(p), surface(c), surface(s)}\\
\texttt{Add:} \nohyphens{clear(s), lifting(h,c)}\\
\texttt{Del:} \nohyphens{at(c,p), available(h), clear(c), on(c,s)}\\
\end{flushleft}

\begin{flushleft}
\underline{Load(c, t, h, p):}\\
\texttt{Pre:} \nohyphens{at(t,p), at(h,p), crate(c), hoist(h), lifting(h,c), locatable(t), locatable(h), locatable(c), object(t), object(h), object(c), surface(c), truck(t)}\\
\texttt{Add:} \nohyphens{available(h), in(c,t)}\\
\texttt{Del:} \nohyphens{lifting(h,c)}\\
\end{flushleft}

\begin{flushleft}
\underline{Drive(from, t, to):}\\
\texttt{Pre:} \nohyphens{at(t,from), locatable(t), object(from), object(t), object(to), place(from), place(to), truck(t)}\\
\texttt{Add:} \nohyphens{at(t,to)}\\
\texttt{Del:} \nohyphens{at(t,from)}\\
\end{flushleft}

\begin{flushleft}
\underline{Unload(c, t, h, p):}\\
\texttt{Pre:} \nohyphens{at(t,p), at(h,p), available(h), crate(c), hoist(h), in(c,t), locatable(t), locatable(h), locatable(c), object(t), object(h), object(c), surface(c), truck(t)}\\
\texttt{Add:} \nohyphens{lifting(h,c)}\\
\texttt{Del:} \nohyphens{available(h), in(c,t)}\\
\end{flushleft}

\begin{flushleft}
\underline{Drop(s, c, p, h):}\\
\texttt{Pre:} \nohyphens{at(s,p), at(h,p), clear(s), crate(c), hoist(h), lifting(h,c), locatable(s), locatable(c), locatable(h), object(s), object(c), object(p), object(h), place(p), surface(s), surface(c)}\\
\texttt{Add:} \nohyphens{at(c,p), available(h), clear(c), on(c,s)}\\
\texttt{Del:} \nohyphens{clear(s), lifting(h,c)}\\
\end{flushleft}

\end{tcolorbox}

\subsubsection{Elevator.} OARU shrinks Up and Down actions from the original domain into one, yet valid, while Board and Depart are fully recognized. Variable p stands for passenger, f1 and f2 are floors.

\begin{tcolorbox}[breakable]
\begin{center}{\bf Elevator}\end{center}
\begin{flushleft}
\underline{Board(p, f1, f2):}\\
\texttt{Pre:} \nohyphens{destin(p,f2), floor(f1), floor(f2), lift-at(f1), origin(p,f1), passenger(p)}\\
\texttt{Add:} \nohyphens{boarded(p)}\\
\texttt{Del:} \nohyphens{}\\
\end{flushleft}

\begin{flushleft}
\underline{Up-Down(p, f1, f2):}\\
\texttt{Pre:} \nohyphens{floor(f1), floor(f2), lift-at(f2), passenger(p)}\\
\texttt{Add:} \nohyphens{lift-at(f1)}\\
\texttt{Del:} \nohyphens{lift-at(f2)}\\
\end{flushleft}

\begin{flushleft}
\underline{Depart(p, f1, f2):}\\
\texttt{Pre:} \nohyphens{boarded(p), destin(p,f2), floor(f1), floor(f2), lift-at(f2), origin(p,f1), passenger(p)}\\
\texttt{Add:} \nohyphens{served(p)}\\
\texttt{Del:} \nohyphens{boarded(p)}\\
\end{flushleft}

\end{tcolorbox}

\subsubsection{Gripper.} This domain is fully recognized without flaws. Variables from, to and r represent rooms, b is a ball and g a gripper. Like in {\em Blocksworld} domain, parameterizing with a gripper makes the domain general for multiple grippers.

\begin{tcolorbox}[breakable]
\begin{center}{\bf Gripper}\end{center}
\begin{flushleft}
\underline{Pick(r, g, b):}\\
\texttt{Pre:} \nohyphens{at(b,r), at-robby(r), ball(b), free(g), gripper(g), room(r)}\\
\texttt{Add:} \nohyphens{carry(b,g)}\\
\texttt{Del:} \nohyphens{at(b,r), free(g)}\\
\end{flushleft}

\begin{flushleft}
\underline{Move(from, to):}\\
\texttt{Pre:} \nohyphens{at-robby(from), room(from), room(to)}\\
\texttt{Add:} \nohyphens{at-robby(to)}\\
\texttt{Del:} \nohyphens{at-robby(from)}\\
\end{flushleft}

\begin{flushleft}
\underline{Drop(g, b, r):}\\
\texttt{Pre:} \nohyphens{at-robby(r), ball(b), carry(b,g), gripper(g), room(r)}\\
\texttt{Add:} \nohyphens{at(b,r), free(g)}\\
\texttt{Del:} \nohyphens{carry(b,g)}\\
\end{flushleft}

\end{tcolorbox}

\subsubsection{Minecraft.} The Recall action is not found because it is never used by goal-oriented agents, always crafting new items that release the agents hand. The other four actions are like in the original domain, but with redundant variables and preconditions. The non-redundant variables stands for agent (a), locations (loc, from and to) and items (i.e. itemN, where N is a possibly empty number).

\begin{tcolorbox}[breakable]
\begin{center}{\bf Minecraft}\end{center}
\begin{flushleft}
\underline{Pick(item, loc, z1):}\\
\texttt{Pre:} \nohyphens{agentat(loc), at(item,loc), moveable(item), static(loc), static(z1)}\\
\texttt{Add:} \nohyphens{inventory(item)}\\
\texttt{Del:} \nohyphens{at(item,loc), static(z1)?}\\
\end{flushleft}

\begin{flushleft}
\underline{Equip(z1, z2, z3, a, z4, item, z5, z6):}\\
\texttt{Pre:} \nohyphens{agent(a), handsfree(a), inventory(item), isgrass(z3), moveable(z1), moveable(z3), moveable(item), static(z2)?, static(z4), static(z5), static(z6)}\\
\texttt{Add:} \nohyphens{equipped(item,a), static(z2)?}\\
\texttt{Del:} \nohyphens{handsfree(a), inventory(item), static(z4)?}\\
\end{flushleft}

\begin{flushleft}
\underline{Craftplank(z1, z2, z3, a, item1, z4, item2, z5, z6):}\\
\texttt{Pre:} \nohyphens{agent(a), at(z1,z3), equipped(item1,a), hypothetical(item2), hypothetical(z5), isgrass(z1), islog(item1), moveable(z1), moveable(item1), moveable(item2), moveable(z5), static(z2)?, static(z3), static(z4), static(z6)}\\
\texttt{Add:} \nohyphens{handsfree(a), inventory(item2), isplanks(item2), static(z2)?}\\
\texttt{Del:} \nohyphens{equipped(item1,a), hypothetical(item2), islog(item1), static(z6)?}\\
\end{flushleft}

\begin{flushleft}
\underline{Move(z, from, to):}\\
\texttt{Pre:} \nohyphens{agentat(from), static(z), static(from), static(to)}\\
\texttt{Add:} \nohyphens{agentat(to)}\\
\texttt{Del:} \nohyphens{agentat(from)}\\
\end{flushleft}

\end{tcolorbox}

\subsubsection{Onearmedgripper.} The same as {\em Gripper} but only with one arm. Actions are correctly recognized but using redundant variables and preconditions. Ball parameter is b1, rooms are r1 and r2, and gripper is g.

\begin{tcolorbox}[breakable]
\begin{center}{\bf Onearmedgripper}\end{center}
\begin{flushleft}
\underline{Pick(r1, b1, z1, z2, z3, g):}\\
\texttt{Pre:} \nohyphens{at(b1,r1), at-robby(r1), ball(b1), ball(z1), ball(z2), free(g), gripper(g), room(r1), room(z3)}\\
\texttt{Add:} \nohyphens{carry(b1,g)}\\
\texttt{Del:} \nohyphens{at(b1,r1), free(g)}\\
\end{flushleft}

\begin{flushleft}
\underline{Move(r1, r2, z1, z2, z3, z4):}\\
\texttt{Pre:} \nohyphens{at-robby(r1), ball(z1), ball(z3), ball(z2), gripper(z4), room(r1), room(r2)}\\
\texttt{Add:} \nohyphens{at-robby(r2)}\\
\texttt{Del:} \nohyphens{at-robby(r1)}\\
\end{flushleft}

\begin{flushleft}
\underline{Drop(r1, z1, b1, z2, z3, g):}\\
\texttt{Pre:} \nohyphens{at-robby(r1), ball(b1), ball(z2), ball(z3), carry(b1,g), gripper(g), room(z1), room(r1)}\\
\texttt{Add:} \nohyphens{at(b1,r1), free(g)}\\
\texttt{Del:} \nohyphens{carry(b1,g)}\\
\end{flushleft}

\end{tcolorbox}

\subsubsection{Rearrangement.} OARU recognizes the original domain but with redundant variables/predicates, and some uncertain predicates that do not affect to the course of the actions. In the preconditions of Move-To-Not-Holding, also detects a casuistic in the observations, where a movable object is in the destination. Location variables are loc, from and to, the agent is the robot variable, and obj is used for pick/placed object variables.

\begin{tcolorbox}[breakable]
\begin{center}{\bf Rearrangement}\end{center}
\begin{flushleft}
\underline{Place(z1, robot, z2, z3, z4, obj):}\\
\texttt{Pre:} \nohyphens{at(z1,z3), holding(obj), isrobot(robot), moveable(z1), moveable(robot), moveable(obj), static(z2)?, static(z3), static(z4)}\\
\texttt{Add:} \nohyphens{handsfree(robot), static(z2)?}\\
\texttt{Del:} \nohyphens{at(z1,z3)?, holding(obj), static(z4)?}\\
\end{flushleft}

\begin{flushleft}
\underline{Move-To-Not-Holding(z, to, robot, from):}\\
\texttt{Pre:} \nohyphens{at(z,to), at(robot,from), handsfree(robot), isrobot(robot), moveable(z), moveable(robot), static(to), static(from)}\\
\texttt{Add:} \nohyphens{at(robot,to)}\\
\texttt{Del:} \nohyphens{at(robot,from)}\\
\end{flushleft}

\begin{flushleft}
\underline{Pick(obj, robot, loc, z):}\\
\texttt{Pre:} \nohyphens{at(obj,loc), at(robot,loc), handsfree(robot), isrobot(robot), moveable(obj), moveable(robot), static(loc), static(z)}\\
\texttt{Add:} \nohyphens{holding(obj)}\\
\texttt{Del:} \nohyphens{handsfree(robot)}\\
\end{flushleft}

\begin{flushleft}
\underline{Move-To-Holding(robot, z, to, from, obj):}\\
\texttt{Pre:} \nohyphens{at(robot,from), at(obj,from), holding(obj), isrobot(robot), moveable(robot), moveable(obj), static(z), static(to), static(from)}\\
\texttt{Add:} \nohyphens{at(robot,to), at(obj,to)}\\
\texttt{Del:} \nohyphens{at(robot,from), at(obj,from)}\\
\end{flushleft}

\end{tcolorbox}

\subsubsection{Sokoban.} In this domain, OARU splits a Push-To-Non-Goal original action into two different actions, Push-From-Goal and Push-To-Clear, where the main difference is that former action pushes a stone from a goal location while the latter action pushes a stone from a clear location to and adjacent clear location. The other two actions are fully recognized. Location parameters are from, to, l1, l2 and l3, directions are d1 and d2, stones are represented with s variable, and the player is p.

\begin{tcolorbox}[breakable]
\begin{center}{\bf Sokoban}\end{center}
\begin{flushleft}
\underline{Push-From-Goal(z1, l1, p, d1, s, l2, z2, z3, l3, d2):}\\
\texttt{Pre:} \nohyphens{at(p,l2), at(s,l3), at-goal(s), clear(l1), clear(z2), is-goal(l3), is-nongoal(z1), is-nongoal(l1), is-nongoal(z2), is-player(p), is-stone(s), location(z1), location(l1), location(l2), location(z2), location(z3), location(l3), move-dir(l1,l3,d2), move-dir(l2,l3,d1), move-dir(l3,l1,d1), move-dir(l3,l2,d2), thing(p), thing(s)}\\
\texttt{Add:} \nohyphens{at(p,l3), at(s,l1), clear(l2)}\\
\texttt{Del:} \nohyphens{at(p,l2), at(s,l3), at-goal(s), clear(l1)}\\
\end{flushleft}

\begin{flushleft}
\underline{Push-To-Clear(l1, d1, l2, p, l3, s, d2):}\\
\texttt{Pre:} \nohyphens{at(p,l2), at(s,l3), clear(l1), is-player(p), is-stone(s), location(l1), location(l2), location(l3), move-dir(l1,l3,d2), move-dir(l2,l3,d1), move-dir(l3,l1,d1), move-dir(l3,l2,d2), thing(p), thing(s)}\\
\texttt{Add:} \nohyphens{at(p,l3), at(s,l1), clear(l2)}\\
\texttt{Del:} \nohyphens{at(p,l2), at(s,l3), clear(l1)}\\
\end{flushleft}

\begin{flushleft}
\underline{Move(to, p, from):}\\
\texttt{Pre:} \nohyphens{at(p,from), clear(to), is-player(p), location(to), location(from), thing(p)}\\
\texttt{Add:} \nohyphens{at(p,to), clear(from)}\\
\texttt{Del:} \nohyphens{at(p,from), clear(to)}\\
\end{flushleft}

\begin{flushleft}
\underline{Push-To-Goal(l1, l2, p, l3, s):}\\
\texttt{Pre:} \nohyphens{at(p,l3), at(s,l2), clear(l1), is-goal(l1), is-nongoal(l2), is-player(p), is-stone(s), location(l1), location(l2), location(l3), thing(p), thing(s)}\\
\texttt{Add:} \nohyphens{at(p,l2), at(s,l1), at-goal(s), clear(l3)}\\
\texttt{Del:} \nohyphens{at(p,l3), at(s,l2), clear(l1)}\\
\end{flushleft}

\end{tcolorbox}

\subsubsection{Travel.}

The domain acquired by OARU ahows an unorthodox structure when compared to the expert's domain. While some of the found actions are reminiscent of the original ones, OARU exploits several quirks of the domain, like the graph structure of the locations, and that some actions are often used together. Thus, OARU combines in a single action predicates that would normally be encoded in different actions (see, for instance, the action we have called \textit{Fly-Short-Drive-And-Visit}).

\begin{tcolorbox}[breakable]
\begin{center}{\bf Travel}\end{center}
\begin{flushleft}
\underline{Fly-Short-Drive-And-Visit(s1, s2, s3, p, c):}\\
\texttt{Pre:} \nohyphens{adjacent(s1,s3), adjacent(s3,s1), at(s2), car(c), caravailable(c), plane(p), planeavailable(p), state(s1), state(s2), state(s3)}\\
\texttt{Add:} \nohyphens{at(s1), visited(s1)}\\
\texttt{Del:} \nohyphens{at(s2), caravailable(c), planeavailable(p)}\\
\end{flushleft}

\begin{flushleft}
\underline{Fly-No-Visit(s1, s2, z1, z2, p):}\\
\texttt{Pre:} \nohyphens{adjacent(s2,z2), adjacent(z1,z2), adjacent(z2,s2), adjacent(z2,z1), at(s1), plane(p), planeavailable(p), state(s1), state(s2), state(s3), visited(s1)}\\
\texttt{Add:} \nohyphens{at(s2)}\\
\texttt{Del:} \nohyphens{at(s1), planeavailable(p)}\\
\end{flushleft}

\begin{flushleft}
\underline{Walk-No-Visit(to, z1, z2, from, z3, z4, z5, z6, z7):}\\
\texttt{Pre:} \nohyphens{adjacent(to,from), adjacent(z1,z2), adjacent(z1,z5), adjacent(z2,z1), adjacent(from,to), adjacent(z3,z4), adjacent(z4,z3), adjacent(z5,z1), adjacent(z5,z6), adjacent(z6,z5), at(from), plane(z7), state(to), state(z1), state(z2), state(from), state(z4), state(z5), visited(from)}\\
\texttt{Add:} \nohyphens{at(to)}\\
\texttt{Del:} \nohyphens{at(from)}\\
\end{flushleft}

\begin{flushleft}
\underline{Drive-And-Visit(c, z1, from, thru, z2, z3, to):}\\
\texttt{Pre:} \nohyphens{adjacent(z1,z2), adjacent(from,thru), adjacent(thru,from), adjacent(z2,z1), adjacent(z3,to), adjacent(to,z3), at(from), car(c), caravailable(c), state(z1), state(from), state(thru), state(z2), state(z3), state(to)}\\
\texttt{Add:} \nohyphens{at(to), visited(to)}\\
\texttt{Del:} \nohyphens{at(from), caravailable(c)}\\
\end{flushleft}

\begin{flushleft}
\underline{Teleport-And-Visit(from, to):}\\
\texttt{Pre:} \nohyphens{at(from), state(from), state(to)}\\
\texttt{Add:} \nohyphens{at(to), visited(to)}\\
\texttt{Del:} \nohyphens{at(from)}\\
\end{flushleft}

\begin{flushleft}
\underline{Fly-And-Visit(z, from, to, p):}\\
\texttt{Pre:} \nohyphens{at(from), plane(p), planeavailable(p), state(z), state(from), state(to)}\\
\texttt{Add:} \nohyphens{at(to), visited(to)}\\
\texttt{Del:} \nohyphens{at(from), planeavailable(p)}\\
\end{flushleft}

\end{tcolorbox}

\end{document}